\documentclass[]{TEAI}
\usepackage{helvet}

\usepackage{amsmath} 
\usepackage{natbib}
\usepackage{graphicx}
\usepackage{subcaption} 

\usepackage[toc,page,header]{appendix}
\usepackage[utf8]{inputenc} 
\usepackage[T1]{fontenc}    
\usepackage{hyperref}       
\usepackage{url}            
\usepackage{xspace}         
\usepackage{booktabs}       
\usepackage{lmodern}        
\usepackage{amsfonts}       
\usepackage{nicefrac}       
\usepackage{microtype}      
\usepackage{wrapfig}

\usepackage{amssymb}  
\usepackage{fontawesome}  
\usepackage{url}  

\usepackage{titletoc}

\usepackage{tikz}  
\usepackage{comment}  
\usepackage{tabularx}  
\usepackage{booktabs}  

\usepackage{minitoc}

\usepackage{booktabs}
\usepackage{array}
\usepackage{etoolbox}

\definecolor{lightblue}{RGB}{200, 230, 255}  
\definecolor{headerblue}{RGB}{150, 200, 255} 

\usepackage{pgfplots}
\usepackage[utf8]{inputenc} 
\usepackage[T1]{fontenc}    
\usepackage{hyperref}       
\usepackage{url}            
\usepackage{booktabs}       
\usepackage{amsfonts}       
\usepackage{nicefrac}       
\usepackage{microtype}      
\usepackage[table,dvipsnames]{xcolor}         
\definecolor{neuripsblue}{rgb}{0.21,0.49,0.74}
\usepackage{graphicx}
\usepackage{float}
\usepackage{comment}
\usepackage{multirow} 
\usepackage{amsmath} 
\usepackage{makecell} 
\usepackage{siunitx}  
\usepackage{tikz}
\usepackage{pgf-pie} 
\usepackage{subcaption}
\usepackage{wrapfig}
\usepackage[export]{adjustbox}
\usepackage{relsize}

\usepackage{ragged2e}      
\usepackage{tabularx}       
\usepackage{array}          
\usepackage{caption}        
\usepackage{enumitem}
\usepackage{pifont}
\usepackage[hang,flushmargin]{footmisc} 

\usepackage{tcolorbox}

\usepackage{tcolorbox}    
\tcbuselibrary{breakable}  
\tcbuselibrary{skins}      

\usepackage{tabularx}
\usepackage{listings}

\newcommand{\sota}[0]{state-of-the-art\xspace}
\newcommand{\ourmethod}{\textsc{Inst-IT}\xspace}
\newcommand{\oursft}{\textsc{Inst-IT}~Dataset\xspace}
\newcommand{\ourbenchmark}[0]{\textsc{Inst-IT}~Bench\xspace}
\newcommand{\fakeparagraph}[1]{\noindent\textbf{#1}}
\newcommand{\eg}{e.g.\ }
\newcommand{\ie}{\textit{i.e.\ }}
\newcommand{\etc}{etc.\ }

\crefname{section}{Sec.}{Secs.}
\Crefname{section}{Section}{Sections}
\Crefname{table}{Table}{Tables}
\crefname{table}{Tab.}{Tabs.}

\newcolumntype{x}[1]{>{\centering\arraybackslash}p{#1pt}}
\newcolumntype{y}[1]{>{\raggedright\arraybackslash}p{#1pt}}
\newlength\savewidth
\newcommand{\tablestyle}[2]{\setlength{\tabcolsep}{#1}\renewcommand{\arraystretch}{#2}\centering\footnotesize}
\newcommand{\up}[1]{{\textcolor{ForestGreen}{~$\uparrow$#1}}}
\newcommand{\down}[1]{{\textcolor{gray}{~$\downarrow$#1}}}

\title{\textsc{Inst-IT}: Boosting Instance Understanding via \\ Explicit Visual Prompt Instruction Tuning}

\author{
    Wujian~Peng\textsuperscript{1,2*},
    Lingchen~Meng\textsuperscript{1*},
    Yitong~Chen\textsuperscript{1,2},  
    Yiweng~Xie\textsuperscript{1},  
    Yang~Liu\textsuperscript{1},  \\
    Tao~Gui\textsuperscript{1,2},  
    Hang~Xu\textsuperscript{3},  
    Xipeng~Qiu\textsuperscript{1,2},  
    Zuxuan~Wu\textsuperscript{1,2,$\dagger$}
    Yu-Gang~Jiang\textsuperscript{1}
}

\affiliation[1]{\mbox{Institute of Trustworthy Embodied AI, Fudan University}}
\affiliation[2]{\mbox{Shanghai Innovation Institute}}
\affiliation[3]{\mbox{Huawei Noah’s Ark Lab}}

\abstract{
Large Multimodal Models~(LMMs) have made significant breakthroughs with the advancement of instruction tuning.
However, while existing models can understand images and videos at a holistic level, they still struggle with instance-level understanding that requires a more fine-grained comprehension and alignment. Instance-level understanding is crucial for LMMs, as it focuses on the specific elements that we are most interested in. Excitingly, existing works find that the \sota LMMs exhibit strong instance understanding capabilities when provided with explicit visual cues. 
Motivated by this, 
we proposed \textbf{\ourmethod}, a solution to enhance LMMs in \uline{\textbf{Inst}}ance understanding via explicit visual prompt \uline{\textbf{I}}nstruction \uline{\textbf{T}}uning for instance guidance.
\ourmethod consists of a benchmark to diagnose multimodal instance-level understanding, a large-scale instruction-tuning dataset, and a continuous instruction-tuning training paradigm to effectively enhance spatial-temporal instance understanding capabilities of existing LMMs. Experimental results show that, enhanced by \ourmethod, our models not only achieve outstanding performance on \ourbenchmark and other instance understanding benchmarks, but also demonstrate significant improvements across various generic image and video understanding benchmarks. This highlights that our method not only boosts instance-level understanding but also strengthens the overall capabilities of generic image and video comprehension.
}

\correspondence{\email{zxwu@fudan.edu.cn}}
\checkdata[Website]{\url{https://inst-it.github.io}}

\begin{document}
\maketitle
\renewcommand{\thefootnote}{}
\footnotetext{$^*$Equal Contribution.\\$^\dagger$Corresponding authors.}
\renewcommand{\thefootnote}{\arabic{footnote}}

\vspace{-1.5em}

\section{Introduction}
\label{sec:intro}
\begin{figure}[t]
  \centering
  \includegraphics[width=\linewidth]{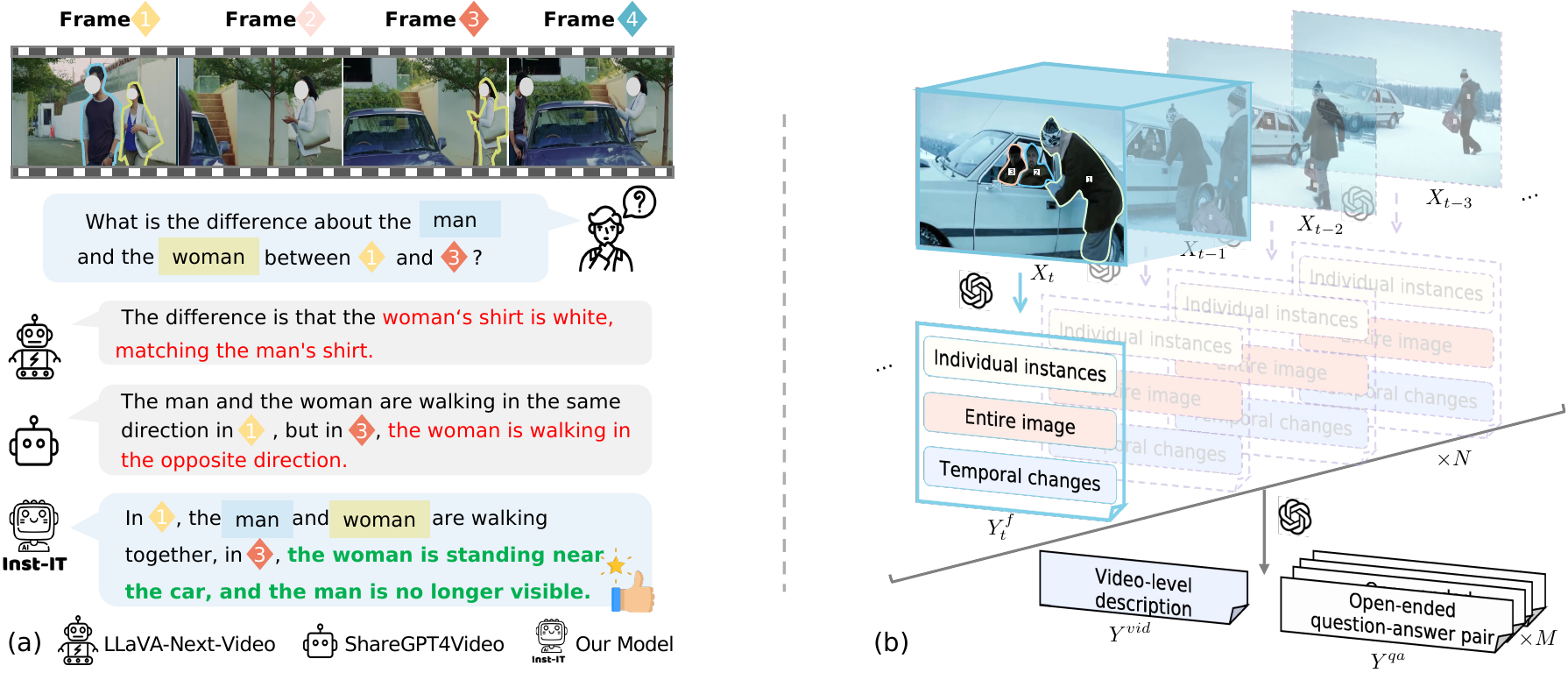}
  \caption{(a) \textbf{LMMs struggle with instance understanding}, failing to capture the nuanced details of instances specified in user queries. (b) \textbf{Our instance-centric data annotation pipeline}, providing multi-level annotations for individual instances in images and videos.}
  \label{fig:teaser}
\end{figure}

Recently, Large Multimodal Models (LMMs) have seen remarkable advancements. A key breakthrough is visual instruction tuning~\cite{llava, instructblip}, enabling models to follow any type of user instructions. This paves the way to building general-purpose multimodal assistants capable of handling a wide range of real-world tasks~\cite{li2024multimodal}. Inspired by this initial work, numerous follow-up studies have emerged in both image-language~\cite{llava1d5, deepstack, minigpt4, sharegpt4v, internvl} and video-language~\cite{video-chatgpt, fu2024objectrelator, videollama, openvclip, Qwen2VL} modeling. 
However, although they can understand images or videos at a holistic level, they still struggle to comprehend instance-specific content that the users are most interested in, as illustrated in \cref{fig:teaser}~(a). 

Instance-level understanding involves comprehending the attributes of individual instances within an image or video, as well as the relationships and interactions between them.
This requires models to exhibit nuanced comprehension and fine-grained alignment. 
Instance understanding has been a long-standing pursuit of the community with extensive efforts devoted to object detection~\cite{faster_rcnn, yolo, fu2024cross, meng2023detection}, instance segmentation~\cite{unet, sam, Meng2023SEGICUT}, and object tracking~\cite{mot, tan2024xtrack}. This capability is essential for real-world applications, where users pay more attention to the instances that they are interested in.
In the era of LMMs, although there have been some attempts in exploring multimodal instance understanding~\cite{regiongpt, vipllava, zhang2023gpt4roi, mgllava, omgllava}, they are primarily limited in the image domain, leaving the videos under-explored. Compared to images, understanding instances in videos is considerably more challenging, as it requires not only capturing their spatial information but also temporal dynamics. Driven by this, \textbf{\textit{we aim to advance the multimodal instance understanding in both images and videos.}}
To this end, we focus on three aspects: instruction-tuning dataset, evaluation benchmark, and training recipe. 

Existing multimodal benchmarks and datasets primarily provide coarse-grained knowledge for images and videos, lacking fine-grained annotations for individual instances. To address this, we introduce an automated pipeline to generate detailed instance-specific annotations.
As illustrated in~\cref{fig:teaser}~(b), we leverage GPT-4o~\cite{openai2024gpt4o} to produce multi-level annotations, including instance-level descriptions, image-level captions, temporal dynamics, video-level summaries, and open-ended question-answer pairs. To fully unleash the capability of GPT-4o for more accurate annotations, we systematically design task prompts and employ set-of-marks visual prompts~\cite{som} to highlight instances in the visual inputs. Powered by this pipeline, we construct \oursft, an instance-grounded multimodal dataset comprising 51k images and 21k videos, 207k image-level captions, 135k temporal dynamics descriptions, 21k video-level captions, and 335k open-ended question-answer pairs.
Furthermore, we carefully design the \ourbenchmark to diagnose the instance-level understanding capabilities of LMMs, and perform rigorous manual verification and refinement to ensure its data quality.

Building on \oursft, we propose a continuous instruction tuning recipe that effectively integrates our instance understanding datasets with general instruction-tuning data. We augment images and videos with visual prompts, and convert the fine-grained annotations from \oursft into instruction tuning format, emphasizing the model's spatiotemporal understanding of individual instances. Experimental results show that our enhanced models achieve strong instance understanding performance not only on \ourbenchmark, but also demonstrate consistent improvements on other instance understanding benchmarks \eg RefCOCOg~\cite{refcocog} and ViP-Bench~\cite{vipllava}. We also investigate the models' general comprehension capabilities on widely used generic benchmarks. The results reveal significant improvements over the baseline, achieving 4.4\% and 13.5\% gains on AI2D~\cite{kembhavi2016diagram} and ChartQA~\cite{masry2022chartqa} image benchmarks, as well as 7.8\% and 11.8\% improvements on Egoschema~\cite{mangalam2023egoschema} and NExT-QA~\cite{xiao2021next} video benchmarks, respectively. This highlights the effectiveness of \ourmethod in boosting instance understanding while strengthening general comprehension in both images and videos.
Our contributions are three-fold:
\begin{enumerate}[leftmargin=2em]
  \item
    We construct the \oursft, the first instance-grounded instruction-tuning dataset that includes both images and videos, featuring explicit instance-level visual prompts and fine-grained annotations grounded on individual instances.
  \item 
    We introduce the \ourbenchmark, a human-verified benchmark specifically designed to evaluate the instance-level understanding capabilities of LMMs on both images and videos.
  \item
  We propose a continuous instruction tuning recipe, which leverages our instance-level dataset alongside general data, effectively enhancing models in instance understanding while consistently improving general comprehension in both images and videos.

\end{enumerate}

\section{Related Work}
\label{sec:related_works}

\fakeparagraph{Large multimodal models.} 
Recently, significant progress has been witnessed in LMMs~\cite{yin2024survey}. BLIP-2~\cite{li2023blip} and Flamingo~\cite{alayrac2022flamingo} leverage visual re-samplers to integrate image features as language inputs by extracting a fixed number of visual tokens. LLaVA~\cite{llava} and its follow-ups~\cite{llava1d5,li2024llava,lin2024vila,mckinzie2025mm1,zhang2024mm1, deepstack, chen2025comp}  achieve remarkable success by connecting vision and language through a simple projection module. Additionally, researchers are extending LMMs’ capabilities to temporal understanding by incorporating multi-frame inputs~\cite{videollava, Qwen2VL, llavavideo} or explicit temporal modules~\cite{li2025llama,he2024ma}
However, existing LMMs struggle with instance-level understanding and often fail to accurately follow instructions to ground specific instances. We emphasize the importance of instance understanding and enhance it through instruction fine-tuning with explicit visual prompts.

\fakeparagraph{Multimodal datasets and benchmarks.}
With the rapid progress in LMMs, numerous instruction-tuning datasets have been developed. LLaVA-Instruct~\cite{llava} leverages object categories, bounding boxes, and image-level captions to generate diverse visual instruction tuning data. Follow-up studies use more powerful models to generate synthetic data~\cite{sharegpt4v,wang2023see,chen2024allava} and improve the annotation pipeline~\cite{li2024densefusion,sharegpt4video,llavavideo}. Simultaneously, various benchmarks are proposed to evaluate LMMs across different aspects~\cite{mmesurvey,li2024survey, li2025benchmark}, such as comprehensive understanding~\cite{li2023seed}, OCR~\cite{masry2022chartqa,mathew2021docvqa,mathew2022infographicvqa,tito2023hierarchical}, temporal understanding~\cite{fu2024video,mangalam2023egoschema,xiao2021next,cai2024temporalbench,liu2024tempcompass,liu2024bench}, and instruction-following~\cite{qian2024mia}. However, they focus more on image or video-level understanding and lack fine-grained emphasis on specific instances. 
We emphasize the importance of instance understanding in both images and videos, and propose the \ourbenchmark to evaluate the instance understanding of LMMs and create the \oursft, providing detailed instance-level annotations to enhance instance understanding.

\fakeparagraph{Multimodal instance understanding.}
Understanding individual instances is a central focus in computer vision community, with key tasks like object detection~\cite{faster_rcnn, yolo, chen2025comprehensive}, instance segmentation~\cite{unet, sam}, and object tracking~\cite{mot, motsurvey, yilmaz2006object}. In the era of LMMs, instance understanding gains increasing attention. SPEC~\cite{spec}, ARO\cite{aro}, and Winoground~\cite{thrush2022winoground} reveal that CLIP~\cite{clip} struggle to understand instances. To address this, KOSMOS-2~\cite{Peng2023Kosmos2GM}, Ferret~\cite{ferret}, GLaMM~\cite{glamm} and Shikra~\cite{chen2023shikra} encode instance information in textual form. In parallel, SoM-LLaVA~\cite{somllava}, RegionGPT~\cite{regiongpt}, GPT4ROI~\cite{zhang2023gpt4roi}, MG-LLaVA~\cite{mgllava}, OMG-LLaVA~\cite{omgllava}, and ViP-LLaVA~\cite{vipllava}, explores the use of visual prompting to guide models in focusing on specific instances. 
SoM-LLaVA~\cite{somllava} and Elysium~\cite{wang2024elysium} are closely related to ours. 
SoM-LLaVA~\cite{somllava} asks models to list the instances in images, finding this effective in enhancing model comprehension. However, it is limited to the image domain. Elysium~\cite{wang2024elysium} focuses on object understanding in videos but employs relatively simplistic instance annotations. 
In contrast, we focus on both images and videos and provide multi-level fine-grained annotations for instances, aiming to advance multimodal models in understanding the spatiotemporal dynamics of individual instances.
\section{\ourmethod}
\label{sec:data_generate}
To address the scarcity of instance-grounded data, we propose an automated pipeline to generate detailed annotations for both images and videos, with a particular emphasis on \textbf{\textit{instances of interest}} (\cref{sec:pipeline}).
Based on this, we build a large-scale instance-grounded multimodal dataset (\cref{sec:oursft}), and carefully design an instance-centric evaluation benchmark (\cref{sec:inst_bench}).
Furthermore, we propose a continuous instruction-tuning recipe (\cref{sec:our_training}) to enhance LMMs in instance understanding.

\subsection{Instance-centric annotation pipeline}
\label{sec:pipeline}
\fakeparagraph{Overview.}
We propose an automated pipeline to generate annotations grounded on individual instances. As in \cref{fig:teaser} (b), we annotate each frame sequentially, aggregate frame-level annotations into a comprehensive video-level description, and generate open-ended question-answer pairs.

\fakeparagraph{Visual prompting.}
Directly processing the raw visual inputs suffers from hallucinations and distraction. To mitigate this issue, we augment the images and videos with visual prompts to highlight the instances. Specifically, we use set-of-marks (SoMs) visual prompt~\cite{som}, which overlays a numerical ID on each instance. We find this method highly effective in guiding GPT-4o to provide annotations focused on individual instances. For more details, please refer to \cref{sec:supp_som}.

\fakeparagraph{Frame-level annotation.}
\label{sec:ann_frame}
We annotate video frames sequentially. At timestamp $t$, we provide GPT-4o with the current frame $X_t$, the previous frame $X_{t\text{-}1}$, and a tailored task prompt $P^f$. We then obtain a frame-level annotation $Y_t^f\text{=}\left( y_t^{ins}, y_t^{img}, y_t^{dif} \right)$ encompassing three aspects, where $y_t^{ins}$ represents the captions for individual instances, $y_t^{img}$ is a caption for the entire image, and $y_t^{dif}$ describes the temporal differences from the previous frame:
\begin{equation}
    Y_t^f = \text{GPT}\left( P^f, X_{t}, X_{t-1} \right).
\end{equation}

\fakeparagraph{Video-level summary.}
\label{sec:ann_video}
After obtaining annotations for each frame, we aggregate them into a caption for the entire video $Y^{vid}$, capturing detailed spatiotemporal information of individual instances: 
\begin{equation}
    Y^{vid} = \text{GPT}\left( P^{vid}, [Y_1^f, Y_2^f, \cdots, Y_N^f] \right),
\end{equation}
where $P^{vid}$ is the task prompt designed for video-level summary and $N$ is the total number of frames.

\fakeparagraph{Open-ended question-answer pairs.}
We also prompt GPT-4o with the task prompt $P^{qa}$ to create $M$ open-ended QA pairs $Y^{qa}\text{=}\{(q_i, a_i)\}_{i=1}^{M}$ focusing on instance understanding:
\label{sec:ann_qa}
\begin{equation}
    Y^{qa} = \text{GPT}\left(P^{qa}, [Y_1^f, Y_2^f, \cdots, Y_N^f]\right).
\end{equation}

\begin{figure}[t]
    \centering
    \includegraphics[width=\linewidth]{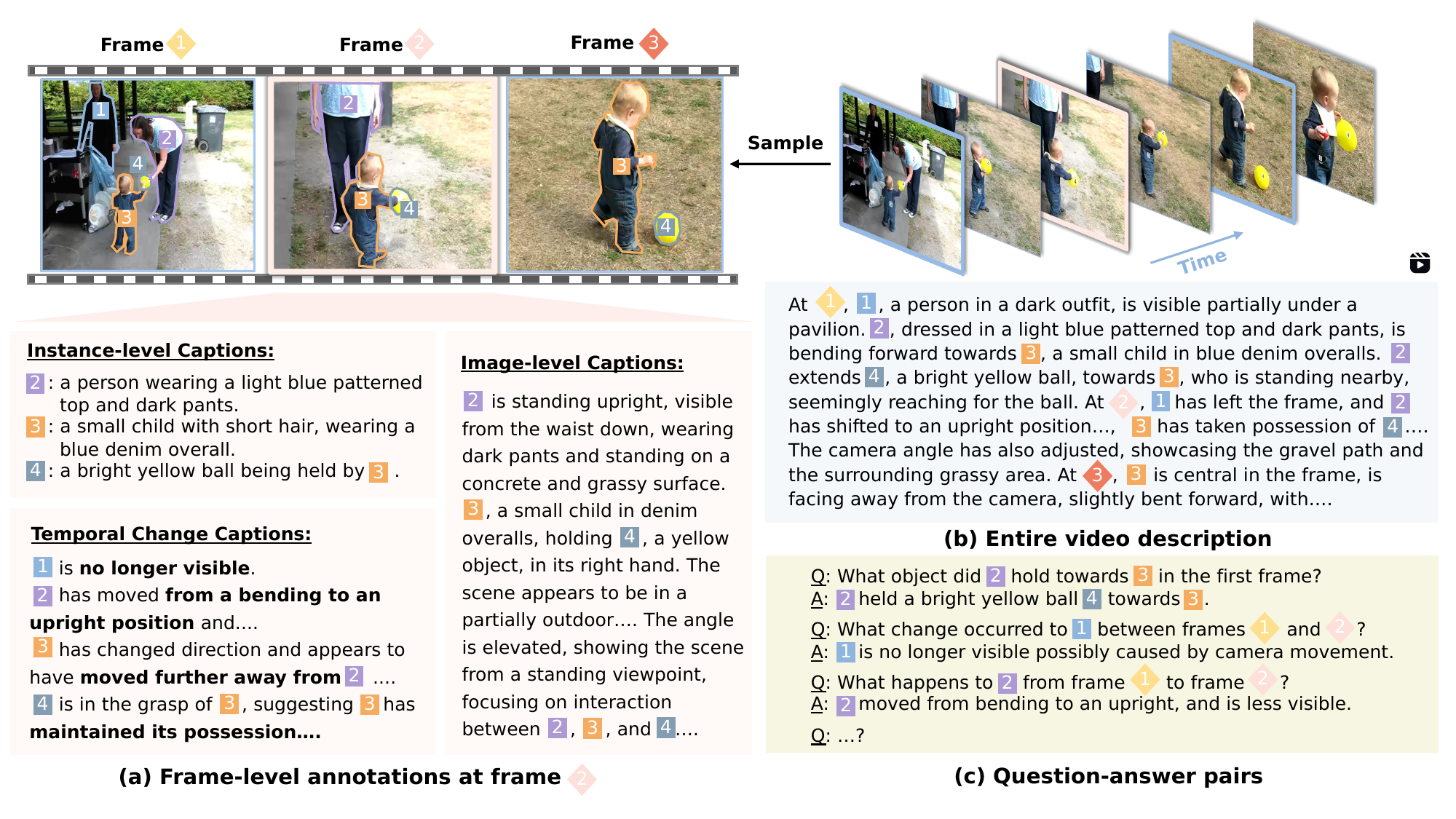}
    \caption{\textbf{Visualization of data structure in \oursft.}
    For each video, we provide (a) $N$ frame-level annotations, (b) a video-level description, and (c) $M$ open-ended question-answer pairs. 
    A complete example data can be found in~\cref{sec:supp_dataset_exp}.
}
    \label{fig:example}
\end{figure}

Following these steps, each video is enriched with multi-granularity annotations that incorporate instance-specific information. As illustrated in~\cref{fig:example}, these annotations include the following aspects:
\begin{itemize}[leftmargin=2em]
    \item \textbf{$N$ frame-level annotations}, each contains detailed descriptions of individual instances, the entire image, and the temporal dynamics between adjacent frames.
    \item \textbf{A comprehensive description} covering the entire video.
    \item \textbf{$M$ open-ended question-answer pairs} that focused on individual instances or their relationships.
\end{itemize}
 Additional information about the design of each task prompt is provided in \cref{sec:supp_prompt_gpt}.

\subsection{\textbf{\oursft}}
\label{sec:oursft}
Instruction tuning plays a crucial role in multimodal training; however, the lack of instance-level datasets hinders the advancement of instance understanding. 
Using the data annotation pipeline described in~\cref{sec:pipeline}, we create a large-scale instruction-tuning dataset, the \oursft. To the best of our knowledge, this is the first instruction fine-tuning dataset that provides multi-level fine-grained annotations centric on individual instances in both images and videos.

\fakeparagraph{Data sources.}
We utilize five video instance segmentation datasets (BRUST~\cite{brust,tao}, UVO~\cite{uvo}, OVIS~\cite{ovis}, LVVIS~\cite{lvvis} and YoutubeVIS-2021~\cite{ytvis}) and two object tracking datasets (BenSMOT~\cite{bensmot}, VidOR~\cite{vidor}) as our video sources, as they provide annotations of instance locations, which is useful in SoM visual prompting~\cite{som}. 
For the image source, we select the SA-1B~\cite{sam} dataset due to its diversity and abundance of instance objects. 
In total, we collect 51k images and 21k videos. More details can be found in \cref{sec:supp_dataset_source}. 

\fakeparagraph{Statistics.}
On average, each video includes one video-level annotation, 7.3 frame-level annotations, and 15.6 open-ended QA pairs. Images are regarded as single-frame videos without temporal changes. In total, \oursft includes 21k videos and 51k images, alongside 21k video-level captions, 207k frame-level captions, 836k instance-level captions, 135k temporal descriptions, and 335k open-ended QA pairs. More statistical analyses are provided in \cref{sec:supp_dataset_statis}. 

\fakeparagraph{Data quality.}
We employ three strategies to ensure the data quality: 
(1) High-quality visual prompts, we use manually annotated labels in segmentation and tracking tasks as SoMs to reduce noise.
(2) Specialized prompt design, we introduce multi-level prompt engineering at the instance, image, two-frame, and video levels to mitigate long-term inconsistencies.
(3) Diversity filtering, we filter out samples with few instances to enhance diversity and domain coverage.
We randomly select 500 data samples and invite 3 volunteers to independently rate each sample with a score ranging from 1 to 5 (higher is better). The mean$_{\pm{\text{std}}}$ of scores and average time spent per sample are in~\cref{tab:human_eval}. The average score is 4.49$_{\pm{\text{0.05}}}$, indicating the satisfactory quality of our data.
We use the maximum score difference ($\text{max}_{\text{diff}}$) among volunteers to assess rating consistency. 49.8\% of samples have $\text{max}_{\text{diff}}$=0, and 78.6\% $\text{max}_{\text{diff}}$ $\leq$ 1, showing high agreements on the ratings of different volunteers. 

\fakeparagraph{Comparison with existing instruction tuning datasets.} 
\cref{tab:data_compare} (left) compares \oursft with other datasets. Prior video datasets, \eg ShareGPT4Video~\cite{sharegpt4video} and LLaVA-Video~\cite{llavavideo}, focus on holistic understanding without instance-level annotations. While VIP-LLaVA~\cite{vipllava} offers instance annotations for images, it does not include any video data.
In contrast, \oursft encompasses both images and videos with multi-level, fine-grained annotations grounded on individual instances.

\subsection{\textbf{\ourbenchmark}}
\label{sec:inst_bench}
Existing benchmarks primarily focus on global understanding, failing to provide more in-depth insights into the instance-level comprehension. We present the \ourbenchmark, specifically designed to diagnose multimodal instance-level understanding in both images and videos.

\fakeparagraph{Construction process.}
To prevent data leakage, we use videos from the test split, ensuring no overlap with \oursft. We apply the pipeline in~\cref{sec:pipeline} to generate 20 open-ended QA pairs for each image and video. Then, we manually review these QA pairs to ensure their accuracy, diversity, and difficulty. Overly simple questions are removed to ensure the remaining ones are instances-centric. We also refine the questions and answers, making necessary rephrasing to ensure correctness. After this rigorous checking process, each sample retains an average of 3.7 carefully polished QA pairs. In addition, we generate three hard negative for each question to construct a multiple-choice QA data with four options. More details are provided in \cref{sec:supp_bench_create}.

\fakeparagraph{Statistics.}
\ourbenchmark comprises 1,000 QA pairs for 338 images and 1,000 QA pairs for 206 videos. Each QA pair supports two evaluation formats, \ie open-ended and multiple-choice.

\fakeparagraph{Metrics.}
For open-ended QAs, we leverage GPT-4o to evaluate the response from a model based on its similarity to the ground-truth answer. For multiple-choice QAs, we calculate the average accuracy across all samples. More details about the metric calculations can be found in \cref{sec:supp_bench_eval}.

\fakeparagraph{Comparison with existing instance understanding benchmarks.} 
\cref{tab:data_compare} (right) highlights the main differences between \ourbenchmark and existing instance understanding benchmarks such as RefCOCO~\cite{refcoco}, RefCOCOg~\cite{refcocog} and ViP-Bench~\cite{vipllava}: (1) its inclusion of evaluation data for both images and videos, pioneered the evaluation in video LMMs; and (2) it supports both open-ended and multiple-choice formats, enabling comprehensive evaluation.

\begin{table}[t]
\centering
\caption{
\textbf{Comparison of \ourmethod with existing datasets and instance understanding benchmarks.}
\textbf{Left:} Instruction tuning datasets. \textbf{Right:} Instance understanding benchmarks.
\textsc{Img} and \textsc{Vid} indicate whether the data contains images or videos, respectively. \textsc{Inst} denotes the availability of instance-level annotations. 
OE and MC indicate open-ended and multiple-choice QA.
}
\label{tab:data_compare}
\begin{minipage}[t]{0.48\linewidth}
  \centering
  \begin{tabular}{lccc}
  \toprule
     & \textsc{Img} & \textsc{Vid} & \textsc{Inst} \\
    \cmidrule{2-4}
    ShareGPT4Video~\cite{sharegpt4video} & & \checkmark &\\
    LLaVA-Video~\cite{llavavideo} & & \checkmark & \\
    ViP-LLaVA-Data~\cite{vipllava} & \checkmark & & \checkmark \\
    \rowcolor{RoyalBlue!5}
    \oursft & \checkmark & \checkmark & \checkmark \\
    \bottomrule
  \end{tabular}
\end{minipage}
\hspace{0.02\linewidth}
\begin{minipage}[t]{0.48\linewidth}
  \centering
  \begin{tabular}{lccc}
  \toprule
      & \textsc{Img} & \textsc{Vid} & \textsc{Task} \\
    \cmidrule{2-4}
    RefCOCO~\cite{refcoco}  & \checkmark &  & caption \\
    RefCOCOg~\cite{refcocog}  & \checkmark &  & caption \\
    ViP-Bench~\cite{vipllava}  & \checkmark & & OE \\
    \rowcolor{RoyalBlue!5}
    \ourbenchmark & \checkmark & \checkmark & OE\&MC \\
    \bottomrule
  \end{tabular}
\end{minipage}
\label{tab:main}
\end{table}

\subsection{Instruction tuning with explicit visual prompt}
\label{sec:our_training}
\fakeparagraph{Architecture.}
We adopt the widely-used LLaVA-NeXT~\cite{llavanext} architecture to evaluate the effectiveness of our \ourmethod. We train our model under an image-video joint training pipeline, where we mix our \oursft with the open-source LLaVA-NeXT-DATA~\cite{llavanext-data}. For single-image samples, we follow the original AnyRes paradigm~\cite{llavanext} to split and encode sub-images according to the aspect ratio. For video and multi-image data, we batch the samples together, encode them, and flatten them into a sequence. Additionally, we apply $2\times2$ spatial pooling to reduce the number of visual tokens in the video inputs. More details are in~\cref{sec:imple}.

\fakeparagraph{Converting \oursft into instruction tuning format.}
\oursft provides annotations at multiple levels of granularity. 
For the instance- and image-level captions in~\cref{fig:example}(a), we use a single frame as input and structure the task as a two-turn dialogue: the model is first prompted to describe all individual instances, followed by a holistic description of the entire scene.
To capture temporal dynamics, we use temporal captions from~\cref{fig:example}(a), asking the model to describe the differences between two consecutive frames.  
The video-level description in~\cref{fig:example}(b) is treated as a captioning task, where the model is instructed to generate a summary based on all video frames.  
For the open-ended QA pairs in~\cref{fig:example}(c), we organize them into a multi-turn conversation, with the model answering one question per turn. 
In total, we construct 243k instruction tuning samples in the form of single-turn and multi-turn dialogues.
All images and video frames are augmented with SoM visual prompts to explicitly provide instance-level guidance.

\begin{table}[t]
\centering
\caption{\textbf{Results on \ourbenchmark.} We conduct evaluations on \ourbenchmark, including \sota open-source image models, video models, and cutting-edge proprietary models. \#IT indicates the number of training samples used during the instruction-tuning stage. N/A indicates that the number is unknown. OE and MC represent open-ended and multiple-choice evaluations, respectively.}
\resizebox{\linewidth}{!}{
\begin{tabular}{lcrcccc}
\toprule
\multicolumn{1}{c}{\multirow{2}{*}{\textbf{Model}}} & \multirow{2}{*}{\textbf{LLM}} & \multirow{2}{*}{\textbf{\#IT}}  & \multicolumn{2}{c}{\textbf{Image}} & \multicolumn{2}{c}{\textbf{Video}} \\ \cmidrule(lr){4-5}\cmidrule(lr){5-7}
 &   & & OE Q\&A  &MC Q\&A  & OE Q\&A  & MC Q\&A  \\ 
\midrule
Random Guess & - & N/A &- & 25.0 & - & 25.0 \\
GPT-4o~\cite{openai2024gpt4o}  & - & N/A & 74.1 & 84.8 & 65.5  & 81.0 \\
Gemini-1.5-flash~\cite{Reid2024Gemini1U}  & - & N/A & 65.3 & 79.5 &  57.9 & 75.8 \\
\multicolumn{7}{c}{\textit{Open-source image models}} \\ 
LLaVA-1.5~\cite{llava1d5}  & Vicuna-7B & 665K &41.6  & 32.1 & - & - \\
ViP-LLaVA~\cite{vipllava}  & Vicuna-7B & $\sim$1.2M &42.1 & 29.2 & - & - \\
SoM-LLaVA~\cite{somllava}  & Vicuna-7B & 695K &45.1 & 40.0 & - & - \\
\rowcolor{RoyalBlue!5}
LLaVA-NeXT~\cite{llavanext}  & Vicuna-7B & 765K &46.0 & 42.4 & - & - \\
\multicolumn{7}{c}{\textit{Open-source video models}} \\ 
LLaVA-NeXT-Video~\cite{llavanext-video}  & Vicuna-7B & 860K &46.5 & 39.5 & 25.8 & 24.8\\
ShareGPT4Video~\cite{sharegpt4video}  & Llama3-8B & $\sim$1.0M &43.2 & 48.7 & 27.8  & 16.1 \\
LLaVA-OV (SI)~\cite{li2024llava}  & Qwen2-7B  & $\sim$7.2M &60.3 &  61.8 & 31.4 & 36.4 \\
LLaVA-OV~\cite{li2024llava}  & Qwen2-7B & $\sim$8.8M &48.0 & 71.7 &  33.2 & 45.6\\
LLaVA-Video~\cite{llavavideo}  & Qwen2-7B & $\sim$7.4M &45.1 & 67.0  & 34.1 & 53.2 \\
InternVL2~\cite{internvl}  & InternLM2.5-7B & N/A &58.6 & 66.5 & 39.8 & 45.5 \\
Qwen2-VL-Instruct~\cite{Qwen2VL}  & Qwen2-72B & N/A &55.5 & 74.7 & 45.5 & 74.6 \\
\multicolumn{7}{c}{\textit{Our models}} \\ 
\rowcolor{RoyalBlue!5}
LLaVA-NeXT-\ourmethod  & Vicuna-7B & 920K &68.6 & 63.0 & 49.3 & 42.1 \\
\rowcolor{RoyalBlue!5}
LLaVA-NeXT-\ourmethod  & Qwen2-7B & 920K &67.9 & 75.3 & 45.7  & 53.3 \\
\bottomrule
\end{tabular}
}
\label{tab:instbench_result}
\end{table}

\section{Experiments}
\label{sec:exps}
\subsection{Implementation details}
\label{sec:imple}
We use LLaVA-NeXT~\cite{llavanext} as our baseline due to its widespread adoption. In the default configuration, Vicuna-1.5-7B~\cite{vicuna2023} serves as the language model with CLIP-ViT-336~\cite{radford2021learning} as the vision encoder. We utilize the AdamW~\cite{loshchilov2017decoupled} with a cosine learning rate schedule for optimization. During the vision-language alignment stage, we use the LCS-558K dataset~\cite{llava1d5}, and for the supervised fine-tuning stage, we leverage the open-source LLaVA-NeXT-DATA~\cite{llavanext-data}. 
For single images, we split the original image into up to 4 sub-images based on its aspect ratio following the AnyRes~\cite{llavanext} approach, and then concatenate the global image with these sub-images. For multiple images and video inputs, we skip the AnyRes procedure and encode every single image. Additionally, we apply $2\times2$ spatial pooling to reduce the number of visual tokens for video inputs. We limit the maximum number of frames to 32 and the context length of LLMs to 6K due to GPU memory constraints.
To enhance instance-level understanding with our \oursft, we combine \oursft with LLaVA-Next-DATA in an additional continuous supervised fine-tuning stage. In this stage, we freeze the first 12 layers of the vision encoder to mitigate potential distribution shifts caused by visually prompted images.
Furthermore, we use Qwen2-7B~\cite{qwen2} with SigLIP-SO400M-384~\cite{siglip} for improved performance in our main experiment, and Qwen2-1.5B with CLIP-ViT-336 for efficiency in our ablation study. 
We use 8$\times$H100 for all experiments. The image-video joint training stage takes approximately 20 hours when using Vicuna-7B as the language model and 24 hours using Qwen2-7B with SigLIP-SO400M-384.

\begin{table}[t]
\centering
\caption{\textbf{Main results on image benchmarks.}}
\tablestyle{2pt}{1.1}
\resizebox{\linewidth}{!}{
\begin{tabular}{lccccccc} 
\toprule
\multirow{2}{*}{\textbf{Method}} & \multirow{2}{*}{\textbf{LLM}} & \multirow{2}{*}{\textbf{Vision Encoder}} & \textbf{AI2D}\cite{kembhavi2016diagram} & \textbf{MMMU}\cite{yue2024mmmu} & \textbf{POPE}\cite{li2023evaluating} & \textbf{GQA}\cite{hudson2019gqa} & \textbf{ChartQA}\cite{masry2022chartqa}\\
& & & (test) & (val) & (test F1) & (val) & (test) \\
\midrule
LLaVA-1.5~\cite{llava1d5} & Vicuna-7B  & CLIP-ViT-Large & 54.8 & 35.3 & 85.9 & 62.0 & 18.2 \\
DeepStack-L~\cite{deepstack} & Vicuna-7B  & CLIP-ViT-Large & - & 35.7 & 86.7 & 63.1 & 21.0 \\
DeepStack-L-HD~\cite{deepstack} & Vicuna-7B  & CLIP-ViT-Large & - & 35.6 & 86.5 & 65.2 & 56.3 \\
VILA~\cite{lin2024vila} & Vicuna-7B  & CLIP-ViT-Large & - & - & 85.5 & 62.3  & - \\
LLaVA-OV (SI)~\cite{li2024llava} & Qwen2-7B  & SigLIP-SO400M & 81.6 & 47.3 & - & - & 78.8 \\
LLaVA-OV~\cite{li2024llava} & Qwen2-7B  & SigLIP-SO400M & 81.4 & 48.8 & - & - & 80.0 \\
Qwen2-VL-Instruct~\cite{Qwen2VL} & Qwen2-7B  & DFN-CLIP-H  & 83.0 & 54.1 & - & - & 83.0 \\
\rowcolor{RoyalBlue!5}
LLaVA-NeXT~\cite{llavanext}~(baseline) & Vicuna-7B  & CLIP-ViT-Large & 66.6 & 35.1 & 86.4 & 64.2 & 54.8 \\
\rowcolor{RoyalBlue!5}
LLaVA-NeXT-\ourmethod~(ours) & Vicuna-7B & CLIP-ViT-Large & 71.0\up{4.4} & 37.4\up{2.3} & 87.2\up{0.8} & 65.9\up{1.7}  & 68.3\up{13.5}\\
\rowcolor{RoyalBlue!5}
LLaVA-NeXT-\ourmethod~(ours) & Qwen2-7B  & SigLIP-SO400 & 78.7\up{12.1} & 42.7\up{7.6} & 87.6\up{0.2} & 65.5\up{1.3} & 72.8\up{18.0}\\
\bottomrule
\end{tabular}
}
\label{tab:main_image}
\end{table}

\begin{table}[t]
\centering
\caption{\textbf{Main results on video benchmarks.} We report the average of MCQA, Y/N and CM in TempCompass for determinism results. $^*$ indicates results reproduced by us.}
\tablestyle{1pt}{1.1}
\resizebox{\linewidth}{!}{
\begin{tabular}{lccccccc} 
\toprule
\multirow{2}{*}{\textbf{Method}} & \multirow{2}{*}{\textbf{LLM}} & \multirow{2}{*}{\textbf{Vision Encoder}} & \textbf{ANetQA}\cite{yu2019activityqa}& \textbf{EgoSchema}\cite{mangalam2023egoschema} & \textbf{NExTQA}\cite{xiao2021next} & \textbf{VideoMME}\cite{fu2024video} & \textbf{TempCompass}\cite{liu2024tempcompass}\\
& & & (oe) & (subset) & (mc) & (w/o subs) & (3 avg) \\
\midrule
DeepStack-L~\cite{deepstack} & Vicuna-7B & CLIP-ViT-Large & 49.3 & 38.4 & 61.0 & - & -\\
Video-ChatGPT~\cite{video-chatgpt} & Vicuna-7B & CLIP-ViT-Large & 35.2 & 47.3& -& -& -\\
VideoLLaMA2~\cite{cheng2024videollama} & Vicuna-7B & CLIP-ViT-Large & 50.2 & - & 51.7 & -& -\\
LLaVA-Next-Video~\cite{llavanext-video} & Vicuna-7B & CLIP-ViT-Large & 53.5 & 43.9 & -& 46.5 & -\\
InternVL2~\cite{internvl} & InternLM-7B & InternViT-300M & - & - & - & 54.0 & -\\
LLaVA-OV~\cite{li2024llava} & Qwen2-7B & SigLIP-SO400M  & 56.6 &  60.1 & 79.4 & 58.2 & 69.4\\
LLaVA-Video~\cite{llavavideo} & Qwen2-7B & SigLIP-SO400M  & 56.5 & 57.3 & 83.2 & 63.3 & -\\
Qwen2-VL-Instruct~\cite{Qwen2VL} & Qwen2-7B & DFN-CLIP-H & - &  66.7 & - & 63.3 & 72.9\\
\rowcolor{RoyalBlue!5}
LLaVA-NeXT~\cite{llavanext}~(baseline) & Vicuna-7B & CLIP-ViT-Large & 53.8 & 50.0$^*$ & 58.4$^*$ & 36.2$^*$ & 56.8$^*$\\
\rowcolor{RoyalBlue!5}
LLaVA-NeXT-\ourmethod~(ours) & Vicuna-7B & CLIP-ViT-Large & 53.7\down{0.1} &  57.8\up{7.8} & 70.2\up{11.8} & 44.3\up{8.1} & 59.8\up{3.0} \\
\rowcolor{RoyalBlue!5}
LLaVA-NeXT-\ourmethod~(ours) & Qwen2-7B & SigLIP-SO400 & 55.2\up{1.4} &  50.4\up{0.4} & 73.0\up{14.6} & 54.0\up{17.8} & 63.9\up{7.1} \\
\bottomrule
\end{tabular}
}
\label{tab:main_video}
\end{table}

\subsection{Main experiments}
\label{sec:main_exp}

\fakeparagraph{Results on \ourbenchmark.} 
We conduct extensive evaluations on \ourbenchmark. The results in \cref{tab:instbench_result} show that with instruction tuning using \oursft, our models achieve a significant improvement of nearly 20\% on average score, validating the effectiveness of \ourmethod. 
Moreover, although ViP-LLaVA~\cite{vipllava} utilizes visual prompts for instruction tuning, it shows minor improvement over its baseline, \ie LLaVA-1.5~\cite{llava1d5}, possibly due to overfitting to its training data. In contrast, our model demonstrates consistent improvements on other instance understanding benchmarks, such as ViP-Bench~\cite{vipllava} and RefCOCOg~\cite{refcocog} (\cref{sec:cross_dataset_eval}), as well as on general-purpose evaluation sets like AI2D and Egoschema (will be discussed in the following sections). This suggests that the model trained with \ourmethod generalizes well to other tasks. 
Qwen2VL-72B does not show substantial improvements over its smaller 7B model, indicating that simply scaling up the model size cannot address the challenges in instance understanding. Similarly, by comparing the amount of instruction tuning data used by each model, we observe that large-scale coarse-grained annotations do not lead to essential improvements either. This highlights the importance of instance-specific annotated data.

\fakeparagraph{Results on generic benchmarks.} 
To evaluate general understanding capabilities, we assess our models on several widely used image and video benchmarks using the LMMs-Eval~\cite{zhang2024lmms}.
To ensure a fair comparison with other models, we primarily report results from their original papers or reproduced results in previous studies. On generic image benchmarks, as shown in \cref{tab:main_image}, \ourmethod consistently outperforms our direct baseline model, \ie LLaVA-NeXT. 
The improvement in AI2D, a benchmark that requires grounding and referring understanding capability, is particularly clear. This suggests that \ourmethod effectively boosts the model in fine-grained understanding. Furthermore, when utilizing a more advanced language model and vision encoder, our method achieves performance comparable to large-scale SFT LMMs, such as LLaVA-OV and Qwen2-VL-Instruct, \textbf{while requiring significantly less computational and data cost. }
For video understanding benchmarks in~\cref{tab:main_video}, \ourmethod significantly outperforms both LLaVA-NeXT and LLaVA-NeXT-Video. These consistent improvements demonstrate that enhancing instance-level understanding through explicit visual prompted instruction tuning is an effective strategy for improving generic spatiotemporal understanding capabilities.

\begin{table}[t]
\centering
\caption{\textbf{Results on ViP-Bench.} We perform evaluation with our \ourmethod models without fine-tuning.}
\tablestyle{2.5pt}{1.1}
\resizebox{\linewidth}{!}{
\begin{tabular}{lccccccc|ccccccc}
\toprule
\multirow{2}{*}{Model} & \multicolumn{7}{c|}{\textbf{Synthesized visual prompts}} & \multicolumn{7}{c}{\textbf{Visual prompts from human}} \\
 &  Rec & OCR & Know  & Math & Rel & Lang & \textbf{All} & Rec & OCR & Know  & Math & Rel & Lang & \textbf{All} \\
 \midrule
\rowcolor[rgb]{0.96,0.96,0.96}
GPT-4V-turbo-detail:high~\cite{achiam2023gpt}  &58.1 &   69.8  &  59.5  &  71.0 &   61.4  &  51.9  &  60.7 & 56.9  &    69.7  &    63.7  &    80.6 &     61.1 &     45.6   &    59.9   \\
\rowcolor[rgb]{0.96,0.96,0.96}
GPT-4V-turbo-detail:low~\cite{achiam2023gpt}  & 53.2 &   50.3 &   55.6  &  67.7 &   57.5 &   57.5  &   52.8 & 51.7  &  50.3  &  59.3  &  60.3  &  55.0 &   43.8    &   51.4  \\
InstructBLIP-7B~\cite{instructblip}  & 36.9 & 16.3 & 34.2 & 22.3 & 26.8 & 7.5 & 31.7 & 38.9 & 17 & 35.4 & 9.7 & 29.3 & 17.5 & 33.3 \\
{Shikra-7B }~\cite{chen2023shikra} &40.2	 & 10.0	 & 28.0 & 3.5	 & 18.9	 & 20.6 & 33.7   & -- & -- & -- & -- & -- & --  & --  \\
{GPT4ROI-7B }~\cite{zhang2023gpt4roi} &35.6 & 	16.7	 & 29.7	 & 9.7 & 	32.5 & 	13.8 & 35.1   & -- & -- & -- & -- & -- & --  & --  \\
Kosmos-2~\cite{peng2023kosmos} & 29.5 & 	14.2 & 	18.5	 & 9.7 & 	7.5	 & 21.9 & 26.9 & -- & -- & -- & -- & -- & --  & --  \\
LLaVA-1.5-7B~\cite{llava1d5}   & 50.8 & 12.4 & 49.2 & 6.5 & 51.8 & 23.8 & 41.6 & 49.1 & 13.0 & 42.9 & 9.7 & 50.0 & 27.5 & 40.2 \\
Qwen-VL-Chat~\cite{bai2023qwen} & 43.0 & 30.4 & 40.2 & 9.7 & 25.7 & 28.7 & 39.2 & 48.7 & 22.1 & 41.2 & 6.5 & 48.2 & 25.0 & 41.7 \\
ViP-LLaVA-7B~\cite{vipllava}  & 54.8 & 18.8 & 52.9 & 9.7 & 53.9 & 42.5 & 45.5 & 55.3 & 17.6 & 45.9 & 8.1 & 44.6 & 33.1 & {46.8} \\
\rowcolor{RoyalBlue!5}
LLaVA-NeXT-\ourmethod-Vicuna-7B & 51.3  & 23.7 & 54.2 & 12.9 & 64.3 &  46.2 &  45.1 & 55.0 & 21.3 & 52.5 & 16.1  & 57.5 & 40.6 & 48.2 \\
\rowcolor{RoyalBlue!5}
LLaVA-NeXT-\ourmethod-Qwen2-7B & 58.9 & 24.5 & 48.5 & 12.9 & 48.2 & 46.3 & \textbf{50.5} & 57.7 & 22.5 & 53.2 & 19.4 & 53.6 & 45.0 & \textbf{49.0}\\
\bottomrule
\end{tabular}
}
\label{tab:regionbench}
\end{table}

\subsection{Evaluation on other instance-understanding benchmarks}
\label{sec:cross_dataset_eval}
To assess whether our model has learned generalizable instance understanding capability, we conducted evaluations on out-of-domain instance understanding benchmarks in \textbf{zero-shot} manner. 

\fakeparagraph{ViP-Bench~\cite{vipllava}} is a region-level understanding benchmark that closely aligns with the objectives of \ourmethod. 
As shown in~\cref{tab:regionbench}, our model exhibits strong generalization performance. In particular, our \ourmethod with Vicuna-7B achieves performance comparable to ViP-LLaVA when using rectangular bounding boxes as visual prompts and even surpasses ViP-LLaVA when employing human-style visual prompts. Notably, our model performs as a generalist under \textbf{zero-shot} evaluation, whereas ViP-LLaVA benefits from in-domain tuning, since it is fine-tuned on the dataset of ViP-Bench. 

\fakeparagraph{RefCOCOg~\cite{refcocog}} is a referring expression comprehension benchmark, with fewer labeling errors than its counterpart RefCOCO~\cite{refcoco}. We evaluate our LLaVA-NeXT-\ourmethod-Vicuna-7B model on this benchmark and observe a clear improvement of 10.8\% over the baseline LLaVA-NeXT-Vicuna-7B (63.0\% vs. 52.2\%). This further confirms that our approach effectively enhances the model in instance understanding, rather than simply overfitting to our \ourmethod data format.

\begin{table}[ht]
\centering
\caption{\textbf{Ablation on data training recipe.} 
L.N. denotes LLaVA-NeXT-Data, while \ourmethod$_{img}$ and \ourmethod$_{vid}$ refer to the image and video subsets of \ourmethod.
\ourmethod-I and \ourmethod-V indicate the multi-choice splits of the image and video part of our \ourbenchmark, respectively.}
\tablestyle{1.2pt}{1.1}
\begin{tabular}{c|l|l|cccc|ccc} 
\toprule
\multirow{2}{*}{\textbf{CL}} & \multirow{2}{*}{\textbf{Tune Enc}} & \multirow{2}{*}{\textbf{Data Combination}} & \textbf{AI2D} & \textbf{POPE} & \textbf{GQA} & \textbf{\ourmethod-I}  & \textbf{Next-QA} & \textbf{VideoMME} & \textbf{\ourmethod-V} \\
& & & (test) & (test F1) & (val) & (mc) & (mc) & (w/o subt) & (mc)\\
\midrule
 & All &  L.N. & 61.1 & 86.9 & 61.4 & 45.3 & 56.6 & 45.7 & 31.3\\
 & All & L.N. \& {\ourmethod$_{vid}$} & 60.7 & 86.1 & 61.2 & 60.7  & 59.7 & 47.1 & 43.0\\
\rowcolor[rgb]{0.96,0.96,0.96}
\checkmark & All  &L.N. \& {\ourmethod$_{vid}$} & 62.3& 86.7 & 62.9 & 61.8 & 62.4 & 46.7 & 44.4 \\
\rowcolor[rgb]{0.96,0.96,0.96}
\checkmark  & None & L.N. \& {\ourmethod$_{vid}$} & 63.1 & 86.9 & 62.5 & 60.2 & 63.2 & 47.2 & 44.3\\
\rowcolor[rgb]{0.96,0.96,0.96}
\checkmark  & Last 12 &L.N. \& {\ourmethod$_{vid}$}& 63.2 & 87.0 & 62.5 & 60.1 & 63.3 & 47.2 & 44.0\\
\rowcolor{RoyalBlue!5}
\checkmark  &None  & L.N. \& {\ourmethod}$_{img+vid}$ & 63.0 & 87.0 & 62.7 & 58.6 & 59.8 & 46.7 & 41.6 \\
\rowcolor{RoyalBlue!5}
\checkmark  &Last 12 & L.N. \& {\ourmethod}$_{img+vid}$ &  63.0 & 87.2 & 62.7 & 59.6 & 64.3 & 46.6 & 43.7\\
\bottomrule
\end{tabular}

\label{tab:training}
\end{table}

\begin{table}[ht]
\centering
\caption{\textbf{Ablation on detailed data combination.} The dataset combination in line \#3 corresponds to the video part of \oursft, while line \#4 represents the complete \oursft by incorporating the image part into line \#3.}
\tablestyle{3pt}{1.1}
\resizebox{\linewidth}{!}{
\begin{tabular}{x{10}|l|ccccc|ccc}
\toprule
\multirow{2}{*}{\#} & \multirow{2}{*}{\textbf{Data Combination}}& \textbf{AI2D} & \textbf{MMMU} & \textbf{POPE} & \textbf{GQA} &\textbf{ \ourmethod-I}& \textbf{Next-QA} & \textbf{VideoMME} & \textbf{\ourmethod-V}\\
& & (test) & (val) & (F1) & (val) & (mc) & (mc) & (w/o subt) & (mc)\\
\midrule
0 & LLaVA-NeXT & 61.1 & 35.9 & 86.9 & 61.4 & 45.3 & 56.6 & 45.7 & 31.3\\
1 & $+$ inst-cap \& img-cap & 63.0 & 35.1 & 86.1 & 62.7 & 58.9 & 62.4 & 46.0 & 33.8\\
2 & $+$ temporal diff & 63.0 & 35.6 & 87.1 & 62.7 & 59.6 & 64.2 & 45.6 & 36.9\\
3 & $+$ video-description \& qa & 63.2 & 34.9 & 87.0 & 62.5 & 60.1 & 63.3 & 47.2 & 44.0\\
4 & $+$ \oursft$_{img}$ & 63.0 & 36.1 & 87.2 & 62.7 & 59.6  & 64.3 & 46.6 & 43.7\\
\bottomrule
\end{tabular}
}
\label{tab:data_comb}
\end{table}

\subsection{Ablation study}
We use Qwen2-1.5B~\cite{qwen2} as the language model and CLIP-ViT-L-336~\cite{clip} as the vision encoder for ablation experiments. We first conduct ablation on the training recipe to investigate how to effectively integrate \oursft with existing academic SFT datasets~\cite{llavanext-data} for a balanced improvement. Next, we perform a detailed analysis of the impact of each component in our \oursft.

\fakeparagraph{Effectiveness of our continuous instruction-tuning paradigm.} As shown in~\cref{tab:training}, directly mixing the video split of \oursft with LLaVA-Next-DATA leads to significant improvements on video benchmarks. However, the performance on generic image understanding slightly declines. We believe this is due to two main reasons: (1) the increased ratio of video data may suppress image understanding; (2) visually prompted images may introduce a distribution shift from natural images. To address these issues, we propose a continuous SFT paradigm based on single-image models and freeze the first 12 layers of the vision encoder to preserve realistic low-level features. Our model achieves balanced performance across both image and video benchmarks with this training approach.

\fakeparagraph{Detailed dataset combination.} 
As illustrated in~\cref{fig:example}, \oursft contains fine-grained annotations at multi-level. 
To investigate the effectiveness of each component in \oursft, we conduct an extensive ablation by progressively adding data components. As shown in~\cref{tab:data_comb}, the instance-level and image-level frame captions are essential for improving instance understanding in images. Meanwhile, temporal differences, along with video-level descriptions and QA, significantly enhance video instance understanding. Finally, incorporating the image component of \oursft enables our model to achieve the most balanced performance across generic image and video understanding benchmarks, as well as our \ourbenchmark.
\section{Conclusion}
\label{sec:conclusion}
Instance understanding that detects, segments, and reasons nuanced relationships among objects has long been the goal of computer vision research, yet limited effort has been made to equip LMMs with such capabilities. We introduced \ourbenchmark,  a carefully curated benchmark for evaluating multimodal instance understanding abilities. Extensive evaluations for a wide range of models demonstrate the limitations of current models for understanding at the instance level. To mitigate this issue, we collected \oursft, the first instruction-tuning dataset with explicit instance-level visual prompts and annotations. Based on \oursft, we proposed \ourmethod, a continuous finetuning framework that excels in instance understanding and general comprehension.

\noindent\textbf{Acknowledgement} This work was supported in part by the National Natural Science Foundation of China (Grant 62472098) and the Science and Technology Commission of Shanghai Municipality (No. 24511103100).

\clearpage

\bibliographystyle{plainnat}
\bibliography{main}

@String(IJCV = {Int. J. Comput. Vis.})

@String(CVPR= {IEEE Conf. Comput. Vis. Pattern Recog.})

@String(ICCV= {Int. Conf. Comput. Vis.})

@String(ECCV= {Eur. Conf. Comput. Vis.})

@String(ICLR = {Int. Conf. Learn. Represent.})

@String(AAAI = {AAAI})

@String(IJCV  = {IJCV})

@String(CVPR  = {CVPR})

@String(ICCV  = {ICCV})

@String(ECCV  = {ECCV})

@String(ICLR  = {ICLR})

@misc{openai2024gpt4o,
 author = {OpenAI},
 title = {{GPT-4o} System Card},
 year = {2024}
}

@article{Reid2024Gemini1U,
  title={Gemini 1.5: Unlocking multimodal understanding across millions of tokens of context},
  author={Machel Reid and Nikolay Savinov and Denis Teplyashin and Dmitry Lepikhin and Timothy P. Lillicrap and Jean-Baptiste Alayrac and Radu Soricut and Angeliki Lazaridou and Orhan Firat and Julian Schrittwieser and Ioannis Antonoglou and Rohan Anil and Sebastian Borgeaud and Andrew M. Dai and Katie Millican and Ethan Dyer and Mia Glaese and Thibault Sottiaux and Benjamin Lee and Fabio Viola and Malcolm Reynolds and Yuanzhong Xu and James Molloy and Jilin Chen and Michael Isard and Paul Barham and Tom Hennigan and Ross McIlroy and Melvin Johnson and Johan Schalkwyk and Eli Collins and Eliza Rutherford and Erica Moreira and Kareem W. Ayoub and Megha Goel and Clemens Meyer and Gregory Thornton and Zhen Yang and Henryk Michalewski and Zaheer Abbas and Nathan Schucher, etc.},
  journal={ArXiv preprint},
  year={2024},

}

@article{internvl,
  title={InternVL: Scaling up Vision Foundation Models and Aligning for Generic Visual-Linguistic Tasks},
  author={Chen, Zhe and Wu, Jiannan and Wang, Wenhai and Su, Weijie and Chen, Guo and Xing, Sen and Zhong, Muyan and Zhang, Qinglong and Zhu, Xizhou and Lu, Lewei and Li, Bin and Luo, Ping and Lu, Tong and Qiao, Yu and Dai, Jifeng},
  journal={arXiv preprint arXiv:2312.14238},
  year={2023}
}

@article{Qwen2VL,
  title={Qwen2-VL: Enhancing Vision-Language Model's Perception of the World at Any Resolution},
  author={Wang, Peng and Bai, Shuai and Tan, Sinan and Wang, Shijie and Fan, Zhihao and Bai, Jinze and Chen, Keqin and Liu, Xuejing and Wang, Jialin and Ge, Wenbin and Fan, Yang and Dang, Kai and Du, Mengfei and Ren, Xuancheng and Men, Rui and Liu, Dayiheng and Zhou, Chang and Zhou, Jingren and Lin, Junyang},
  journal={arXiv preprint arXiv:2409.12191},
  year={2024}
}

@article{llavavideo,
  title={Video Instruction Tuning With Synthetic Data},
  author={Yuanhan Zhang and Jinming Wu and Wei Li and Bo Li and Zejun Ma and Ziwei Liu and Chunyuan Li},
  journal={arXiv preprint arXiv:2410.02713},
  year={2024}
}

@inproceedings{sharegpt4video,
title={ShareGPT4Video: Improving Video Understanding and Generation with Better Captions},
author={Chen, Lin and Wei, Xilin and Li, Jinsong and Dong, Xiaoyi and Zhang, Pan and Zang, Yuhang and Chen, Zehui and Duan, Haodong and Lin, Bin and Tang, Zhenyu and Yuan, Li and Qiao, Yu and Lin, Dahua and Zhao, Feng and Wang, Jiaqi},
booktitle={NeurIPS},
year={2024}
}

@inproceedings{brust,
  title={Burst: A benchmark for unifying object recognition, segmentation and tracking in video},
  author={Athar, Ali and Luiten, Jonathon and Voigtlaender, Paul and Khurana, Tarasha and Dave, Achal and Leibe, Bastian and Ramanan, Deva},
  booktitle={WACV},
  year={2023}
}

@inproceedings{sam,
  title={Segment anything},
  author={Kirillov, Alexander and Mintun, Eric and Ravi, Nikhila and Mao, Hanzi and Rolland, Chloe and Gustafson, Laura and Xiao, Tete and Whitehead, Spencer and Berg, Alexander C and Lo, Wan-Yen and others},
  booktitle={ICCV},
  year={2023}
}

@article{tan2024xtrack,
  title={XTrack: Multimodal Training Boosts RGB-X Video Object Trackers},
  author={Tan, Yuedong and Wu, Zongwei and Fu, Yuqian and Zhou, Zhuyun and Sun, Guolei and Zamfi, Eduard and Ma, Chao and Paudel, Danda Pani and Van Gool, Luc and Timofte, Radu},
  journal={ArXiv},
  year={2024}
}

@article{ravi2024sam,
  title={Sam 2: Segment anything in images and videos},
  author={Ravi, Nikhila and Gabeur, Valentin and Hu, Yuan-Ting and Hu, Ronghang and Ryali, Chaitanya and Ma, Tengyu and Khedr, Haitham and R{\"a}dle, Roman and Rolland, Chloe and Gustafson, Laura and others},
  journal={arXiv preprint arXiv:2408.00714},
  year={2024}
}

@inproceedings{bensmot,
  title={Beyond MOT: Semantic Multi-Object Tracking},
  author={Li, Yunhao and Li, Qin and Wang, Hao and Ma, Xue and Yao, Jiali and Dong, Shaohua and Fan, Heng and Zhang, Libo},
  booktitle={ECCV},
  year={2024},
}

@article{ovis,
  title={Occluded video instance segmentation: A benchmark},
  author={Qi, Jiyang and Gao, Yan and Hu, Yao and Wang, Xinggang and Liu, Xiaoyu and Bai, Xiang and Belongie, Serge and Yuille, Alan and Torr, Philip HS and Bai, Song},
  journal={IJCV},
  year={2022},
}

@inproceedings{tao,
  title={Tao: A large-scale benchmark for tracking any object},
  author={Dave, Achal and Khurana, Tarasha and Tokmakov, Pavel and Schmid, Cordelia and Ramanan, Deva},
  booktitle={ECCV},
  year={2020},
}

@inproceedings{uvo,
  title={Unidentified video objects: A benchmark for dense, open-world segmentation},
  author={Wang, Weiyao and Feiszli, Matt and Wang, Heng and Tran, Du},
  booktitle={ICCV},
  year={2021}
}

@article{lvvis,
  title={OV-VIS: Open-Vocabulary Video Instance Segmentation},
  author={Wang, Haochen and Yan, Cilin and Chen, Keyan and Jiang, Xiaolong and Tang, Xu and Hu, Yao and Kang, Guoliang and Xie, Weidi and Gavves, Efstratios},
  journal={IJCV},
  year={2024},
}

@inproceedings{ytvis,
  title={Video instance segmentation},
  author={Yang, Linjie and Fan, Yuchen and Xu, Ning},
  booktitle={ICCV},
  year={2019}
}

@inproceedings{vidor,
    title={Annotating Objects and Relations in User-Generated Videos},
    author={Shang, Xindi and Di, Donglin and Xiao, Junbin and Cao, Yu and Yang, Xun and Chua, Tat-Seng},
    booktitle={ICMR},
    year={2019},
}

@article{som,
  title={Set-of-mark prompting unleashes extraordinary visual grounding in gpt-4v},
  author={Yang, Jianwei and Zhang, Hao and Li, Feng and Zou, Xueyan and Li, Chunyuan and Gao, Jianfeng},
  journal={arXiv preprint arXiv:2310.11441},
  year={2023}
}

@inproceedings{somllava,
  title={List Items One by One: A New Data Source and Learning Paradigm for Multimodal LLMs},
  author={Yan, An and Yang, Zhengyuan and Wu, Junda and Zhu, Wanrong and Yang, Jianwei and Li, Linjie and Lin, Kevin and Wang, Jianfeng and McAuley, Julian and Gao, Jianfeng and others},
  booktitle={COLM},
  year={2024}
}

@inproceedings{llava,
  title={Visual instruction tuning},
  author={Liu, Haotian and Li, Chunyuan and Wu, Qingyang and Lee, Yong Jae},
  booktitle={NeurIPS},
  year={2023}
}

@article{li2024multimodal,
  title={Multimodal foundation models: From specialists to general-purpose assistants},
  author={Li, Chunyuan and Gan, Zhe and Yang, Zhengyuan and Yang, Jianwei and Li, Linjie and Wang, Lijuan and Gao, Jianfeng and others},
  journal={FTCGV},
  year={2024},
}

@inproceedings{llava1d5,
  title={Improved baselines with visual instruction tuning},
  author={Liu, Haotian and Li, Chunyuan and Li, Yuheng and Lee, Yong Jae},
  booktitle={CVPR},
  year={2024}
}

@misc{llavanext-video,
  title={LLaVA-NeXT: A Strong Zero-shot Video Understanding Model},
  url={https://llava-vl.github.io/blog/2024-04-30-llava-next-video/},
  author={Zhang, Yuanhan and Li, Bo and Liu, haotian and Lee, Yong jae and Gui, Liangke and Fu, Di and Feng, Jiashi and Liu, Ziwei and Li, Chunyuan},
  year={2024}
}

@misc{llavanext,
    title={LLaVA-NeXT: Improved reasoning, OCR, and world knowledge},
    url={https://llava-vl.github.io/blog/2024-01-30-llava-next/},
    author={Liu, Haotian and Li, Chunyuan and Li, Yuheng and Li, Bo and Zhang, Yuanhan and Shen, Sheng and Lee, Yong Jae},
    year={2024}
}

@inproceedings{sharegpt4v,
  title={Sharegpt4v: Improving large multi-modal models with better captions},
  author={Chen, Lin and Li, Jinsong and Dong, Xiaoyi and Zhang, Pan and He, Conghui and Wang, Jiaqi and Zhao, Feng and Lin, Dahua},
  booktitle={ECCV},
  year={2024}
}

@inproceedings{instructblip,
  title={InstructBLIP: Towards General-purpose Vision-Language Models with Instruction Tuning},
  author={Wenliang Dai and Junnan Li and Dongxu Li and Anthony Meng Huat Tiong and Junqi Zhao and Weisheng Wang and Boyang Albert Li and Pascale Fung and Steven C. H. Hoi},
  booktitle={NeurIPS},
  year={2023},
}

@inproceedings{deepstack,
  title={DeepStack: Deeply Stacking Visual Tokens is Surprisingly Simple and Effective for LMMs},
  author={Meng, Lingchen and Yang, Jianwei and Tian, Rui and Dai, Xiyang and Wu, Zuxuan and Gao, Jianfeng and Jiang, Yu-Gang},
  booktitle={NeurIPS},
  year={2024}
}

@article{minigpt4,
  title={MiniGPT-4: Enhancing Vision-Language Understanding with Advanced Large Language Models},
  author={Zhu, Deyao and Chen, Jun and Shen, Xiaoqian and Li, Xiang and Elhoseiny, Mohamed},
  journal={ArXiv},
  year={2023}
}

@inproceedings{video-chatgpt,
  title={Video-chatgpt: Towards detailed video understanding via large vision and language models},
  author={Maaz, Muhammad and Rasheed, Hanoona and Khan, Salman and Khan, Fahad Shahbaz},
  booktitle={ACL},
  year={2024}
}

@inproceedings{videollama,
  title={Video-LLaMA: An Instruction-tuned Audio-Visual Language Model for Video Understanding},
  author={Hang Zhang and Xin Li and Lidong Bing},
  booktitle={EMNLP},
  year={2023}
}

@inproceedings{videollava,
  title={Video-LLaVA: Learning United Visual Representation by Alignment Before Projection},
  author={Lin, Bin and Zhu, Bin and Ye, Yang and Ning, Munan and Jin, Peng and Yuan, Li},
  booktitle={EMNLP},
  year={2024}
}

@inproceedings{regiongpt,
  title={RegionGPT: Towards Region Understanding Vision Language Model},
  author={Qiushan Guo and Shalini De Mello and Hongxu Yin and Wonmin Byeon and Ka Chun Cheung and Yizhou Yu and Ping Luo and Sifei Liu},
  booktitle={CVPR},
  year={2024},
}

@article{mgllava,
  title={MG-LLaVA: Towards Multi-Granularity Visual Instruction Tuning},
  author={Zhao, Xiangyu and Li, Xiangtai and Duan, Haodong and Huang, Haian and Li, Yining and Chen, Kai and Yang, Hua},
  journal={ArXiv},
  year={2024}
}

@inproceedings{vipllava,
author={Cai, Mu and Liu, Haotian and Mustikovela,  Siva Karthik and Meyer, Gregory P. and Chai, Yuning and Park, Dennis and Lee, Yong Jae},
title={Making Large Multimodal Models Understand Arbitrary Visual Prompts},
booktitle={CVPR},
year={2024}
}

@inproceedings{omgllava,
  title={OMG-LLaVA: Bridging Image-level, Object-level, Pixel-level Reasoning and Understanding},
  author={Zhang, Tao and Li, Xiangtai and Fei, Hao and Yuan, Haobo and Wu, Shengqiong and Ji, Shunping and Chen, Change Loy and Yan, Shuicheng},
  booktitle={NeurIPS},
  year={2024}
}

@inproceedings{spec,
  title={Synthesize Diagnose and Optimize: Towards Fine-Grained Vision-Language Understanding},
  author={Peng, Wujian and Xie, Sicheng and You, Zuyao and Lan, Shiyi and Wu, Zuxuan},
  booktitle={CVPR},
  year={2024}
}

@inproceedings{aro,
  title={When and why vision-language models behave like bags-of-words, and what to do about it?},
  author={Mert Yuksekgonul and Federico Bianchi and Pratyusha Kalluri and Dan Jurafsky and James Y. Zou},
  booktitle={ICLR},
  year={2023}
}

@inproceedings{li2023blip,
  title={Blip-2: Bootstrapping language-image pre-training with frozen image encoders and large language models},
  author={Li, Junnan and Li, Dongxu and Savarese, Silvio and Hoi, Steven},
  booktitle={ICML},
  year={2023},
}

@inproceedings{alayrac2022flamingo,
  title={Flamingo: a visual language model for few-shot learning},
  author={Alayrac, Jean-Baptiste and Donahue, Jeff and Luc, Pauline and Miech, Antoine and Barr, Iain and Hasson, Yana and Lenc, Karel and Mensch, Arthur and Millican, Katherine and Reynolds, Malcolm and others},
  booktitle={NeurIPS},
  year={2022}
}

@article{li2024llava,
  title={Llava-onevision: Easy visual task transfer},
  author={Li, Bo and Zhang, Yuanhan and Guo, Dong and Zhang, Renrui and Li, Feng and Zhang, Hao and Zhang, Kaichen and Li, Yanwei and Liu, Ziwei and Li, Chunyuan},
  journal={arXiv preprint arXiv:2408.03326},
  year={2024}
}

@inproceedings{lin2024vila,
  title={Vila: On pre-training for visual language models},
  author={Lin, Ji and Yin, Hongxu and Ping, Wei and Molchanov, Pavlo and Shoeybi, Mohammad and Han, Song},
  booktitle={CVPR},
  year={2024}
}

@inproceedings{li2025llama,
  title={Llama-vid: An image is worth 2 tokens in large language models},
  author={Li, Yanwei and Wang, Chengyao and Jia, Jiaya},
  booktitle={ECCV},
  year={2024},
}

@inproceedings{he2024ma,
  title={Ma-lmm: Memory-augmented large multimodal model for long-term video understanding},
  author={He, Bo and Li, Hengduo and Jang, Young Kyun and Jia, Menglin and Cao, Xuefei and Shah, Ashish and Shrivastava, Abhinav and Lim, Ser-Nam},
  booktitle={CVPR},
  year={2024}
}

@inproceedings{hudson2019gqa,
  title={Gqa: A new dataset for real-world visual reasoning and compositional question answering},
  author={Hudson, Drew A and Manning, Christopher D},
  booktitle={CVPR},
  year={2019}
}

@article{masry2022chartqa,
  title={Chartqa: A benchmark for question answering about charts with visual and logical reasoning},
  author={Masry, Ahmed and Long, Do Xuan and Tan, Jia Qing and Joty, Shafiq and Hoque, Enamul},
  journal={arXiv preprint arXiv:2203.10244},
  year={2022}
}

@inproceedings{mathew2021docvqa,
  title={Docvqa: A dataset for vqa on document images},
  author={Mathew, Minesh and Karatzas, Dimosthenis and Jawahar, CV},
  booktitle={WACV},
  year={2021}
}

@inproceedings{mathew2022infographicvqa,
  title={Infographicvqa},
  author={Mathew, Minesh and Bagal, Viraj and Tito, Rub{\`e}n and Karatzas, Dimosthenis and Valveny, Ernest and Jawahar, CV},
  booktitle={WACV},
  year={2022}
}

@article{tito2023hierarchical,
  title={Hierarchical multimodal transformers for Multipage DocVQA},
  author={Tito, Rub{\`e}n and Karatzas, Dimosthenis and Valveny, Ernest},
  journal={Pattern Recognition},
  year={2023},
  publisher={Elsevier}
}

@article{fu2024video,
  title={Video-mme: The first-ever comprehensive evaluation benchmark of multi-modal llms in video analysis},
  author={Fu, Chaoyou and Dai, Yuhan and Luo, Yongdong and Li, Lei and Ren, Shuhuai and Zhang, Renrui and Wang, Zihan and Zhou, Chenyu and Shen, Yunhang and Zhang, Mengdan and others},
  journal={arXiv preprint arXiv:2405.21075},
  year={2024}
}

@inproceedings{mangalam2023egoschema,
  title={Egoschema: A diagnostic benchmark for very long-form video language understanding},
  author={Mangalam, Karttikeya and Akshulakov, Raiymbek and Malik, Jitendra},
  booktitle={NeurIPS},
  year={2023}
}

@inproceedings{xiao2021next,
  title={Next-qa: Next phase of question-answering to explaining temporal actions},
  author={Xiao, Junbin and Shang, Xindi and Yao, Angela and Chua, Tat-Seng},
  booktitle={CVPR},
  year={2021}
}

@article{li2023evaluating,
  title={Evaluating object hallucination in large vision-language models},
  author={Li, Yifan and Du, Yifan and Zhou, Kun and Wang, Jinpeng and Zhao, Wayne Xin and Wen, Ji-Rong},
  journal={arXiv preprint arXiv:2305.10355},
  year={2023}
}

@misc{vicuna2023,
    title = {Vicuna: An Open-Source Chatbot Impressing GPT-4 with 90\%* ChatGPT Quality},
    url = {https://lmsys.org/blog/2023-03-30-vicuna/},
    author = {Chiang, Wei-Lin and Li, Zhuohan and Lin, Zi and Sheng, Ying and Wu, Zhanghao and Zhang, Hao and Zheng, Lianmin and Zhuang, Siyuan and Zhuang, Yonghao and Gonzalez, Joseph E. and Stoica, Ion and Xing, Eric P.},
    month = {March},
    year = {2023}
}

@article{loshchilov2017decoupled,
  title={Decoupled weight decay regularization},
  author={Loshchilov, I},
  journal={arXiv preprint arXiv:1711.05101},
  year={2017}
}

@misc{llavanext-data,
  title={LLaVA-NeXT-Data},
  author={lmms-lab},
  url={https://llava-vl.github.io/blog/2024-01-30-llava-next/},
  year={2024}
}

@inproceedings{radford2021learning,
  title={Learning transferable visual models from natural language supervision},
  author={Radford, Alec and Kim, Jong Wook and Hallacy, Chris and Ramesh, Aditya and Goh, Gabriel and Agarwal, Sandhini and Sastry, Girish and Askell, Amanda and Mishkin, Pamela and Clark, Jack and others},
  booktitle={ICML},
  year={2021},
}

@article{cheng2024videollama,
  title={VideoLLaMA 2: Advancing Spatial-Temporal Modeling and Audio Understanding in Video-LLMs},
  author={Cheng, Zesen and Leng, Sicong and Zhang, Hang and Xin, Yifei and Li, Xin and Chen, Guanzheng and Zhu, Yongxin and Zhang, Wenqi and Luo, Ziyang and Zhao, Deli and others},
  journal={arXiv preprint arXiv:2406.07476},
  year={2024}
}

@article{wang2023see,
  title={To see is to believe: Prompting gpt-4v for better visual instruction tuning},
  author={Wang, Junke and Meng, Lingchen and Weng, Zejia and He, Bo and Wu, Zuxuan and Jiang, Yu-Gang},
  journal={arXiv preprint arXiv:2311.07574},
  year={2023}
}

@misc{chen2024allava,
      title={ALLaVA: Harnessing GPT4V-synthesized Data for A Lite Vision-Language Model}, 
      author={Guiming Hardy Chen and Shunian Chen and Ruifei Zhang and Junying Chen and Xiangbo Wu and Zhiyi Zhang and Zhihong Chen and Jianquan Li and Xiang Wan and Benyou Wang},
      year={2024},
      eprint={2402.11684},
      archivePrefix={arXiv},
      primaryClass={cs.CL}
}

@article{achiam2023gpt,
  title={Gpt-4 technical report},
  author={Achiam, Josh and Adler, Steven and Agarwal, Sandhini and Ahmad, Lama and Akkaya, Ilge and Aleman, Florencia Leoni and Almeida, Diogo and Altenschmidt, Janko and Altman, Sam and Anadkat, Shyamal and others},
  journal={arXiv preprint arXiv:2303.08774},
  year={2023}
}

@article{chen2023shikra,
  title={Shikra: Unleashing multimodal llm's referential dialogue magic},
  author={Chen, Keqin and Zhang, Zhao and Zeng, Weili and Zhang, Richong and Zhu, Feng and Zhao, Rui},
  journal={arXiv preprint arXiv:2306.15195},
  year={2023}
}

@article{zhang2023gpt4roi,
  title={Gpt4roi: Instruction tuning large language model on region-of-interest},
  author={Zhang, Shilong and Sun, Peize and Chen, Shoufa and Xiao, Min and Shao, Wenqi and Zhang, Wenwei and Liu, Yu and Chen, Kai and Luo, Ping},
  journal={arXiv preprint arXiv:2307.03601},
  year={2023}
}

@article{bai2023qwen,
  title={Qwen-vl: A versatile vision-language model for understanding, localization, text reading, and beyond},
  author={Bai, Jinze and Bai, Shuai and Yang, Shusheng and Wang, Shijie and Tan, Sinan and Wang, Peng and Lin, Junyang and Zhou, Chang and Zhou, Jingren},
  journal={arXiv preprint arXiv:2308.12966},
  volume={1},
  number={2},
  pages={3},
  year={2023}
}

@article{peng2023kosmos,
  title={Kosmos-2: Grounding multimodal large language models to the world},
  author={Peng, Zhiliang and Wang, Wenhui and Dong, Li and Hao, Yaru and Huang, Shaohan and Ma, Shuming and Wei, Furu},
  journal={arXiv preprint arXiv:2306.14824},
  year={2023}
}

@inproceedings{thrush2022winoground,
  title={Winoground: Probing vision and language models for visio-linguistic compositionality},
  author={Thrush, Tristan and Jiang, Ryan and Bartolo, Max and Singh, Amanpreet and Williams, Adina and Kiela, Douwe and Ross, Candace},
  booktitle={CVPR},
  year={2022}
}

@article{faster_rcnn,
  title={Faster R-CNN: Towards Real-Time Object Detection with Region Proposal Networks},
  author={Shaoqing Ren and Kaiming He and Ross B. Girshick and Jian Sun},
  journal={TPAMI},
  year={2015}
}

@article{unet,
  title={U-Net: Convolutional Networks for Biomedical Image Segmentation},
  author={Olaf Ronneberger and Philipp Fischer and Thomas Brox},
  journal={ArXiv},
  year={2015},
}

@article{mot,
  title={MOTChallenge: A Benchmark for Single-Camera Multiple Target Tracking},
  author={Patrick Dendorfer and Aljosa Osep and Anton Milan and Konrad Schindler and Daniel Cremers and Ian D. Reid and Stefan Roth and Laura Leal-Taix{\'e}},
  journal={IJCV},
  year={2020},
  volume={129},
}

@inproceedings{clip,
  title={Learning Transferable Visual Models From Natural Language Supervision},
  author={Alec Radford and Jong Wook Kim and Chris Hallacy and Aditya Ramesh and Gabriel Goh and Sandhini Agarwal and Girish Sastry and Amanda Askell and Pamela Mishkin and Jack Clark and Gretchen Krueger and Ilya Sutskever},
  booktitle={ICML},
  year={2021}
}

@article{Peng2023Kosmos2GM,
  title={Kosmos-2: Grounding Multimodal Large Language Models to the World},
  author={Zhiliang Peng and Wenhui Wang and Li Dong and Yaru Hao and Shaohan Huang and Shuming Ma and Furu Wei},
  journal={ArXiv},
  year={2023},
  volume={abs/2306.14824},
  url={https://api.semanticscholar.org/CorpusID:259262263}
}

@article{qwen2,
  title={Qwen2 Technical Report},
  author={An Yang and Baosong Yang and Binyuan Hui and Bo Zheng and Bowen Yu and Chang Zhou and Chengpeng Li and Chengyuan Li and Dayiheng Liu and Fei Huang and Guanting Dong and Haoran Wei and Huan Lin and Jialong Tang and Jialin Wang and Jian Yang and Jianhong Tu and Jianwei Zhang and Jianxin Ma and Jin Xu and Jingren Zhou and Jinze Bai and Jinzheng He and Junyang Lin and Kai Dang and Keming Lu and Ke-Yang Chen and Kexin Yang and Mei Li and Min Xue and Na Ni and Pei Zhang and Peng Wang and Ru Peng and Rui Men and Ruize Gao and Runji Lin and Shijie Wang and Shuai Bai and Sinan Tan and Tianhang Zhu and Tianhao Li and Tianyu Liu and Wenbin Ge and Xiaodong Deng and Xiaohuan Zhou and Xingzhang Ren and Xinyu Zhang and Xipin Wei and Xuancheng Ren and Yang Fan and Yang Yao and Yichang Zhang and Yunyang Wan and Yunfei Chu and Zeyu Cui and Zhenru Zhang and Zhi-Wei Fan},
  journal={ArXiv},
  year={2024},
  volume={abs/2407.10671},
}

@inproceedings{siglip,
  title={Sigmoid Loss for Language Image Pre-Training},
  author={Xiaohua Zhai and Basil Mustafa and Alexander Kolesnikov and Lucas Beyer},
  booktitle={ICCV},
  year={2023},
}

@article{zhang2024lmms,
  title={Lmms-eval: Reality check on the evaluation of large multimodal models},
  author={Zhang, Kaichen and Li, Bo and Zhang, Peiyuan and Pu, Fanyi and Cahyono, Joshua Adrian and Hu, Kairui and Liu, Shuai and Zhang, Yuanhan and Yang, Jingkang and Li, Chunyuan and others},
  journal={arXiv preprint arXiv:2407.12772},
  year={2024}
}

@article{li2023seed,
  title={Seed-bench: Benchmarking multimodal llms with generative comprehension},
  author={Li, Bohao and Wang, Rui and Wang, Guangzhi and Ge, Yuying and Ge, Yixiao and Shan, Ying},
  journal={arXiv preprint arXiv:2307.16125},
  year={2023}
}

@article{cai2024temporalbench,
  title={TemporalBench: Benchmarking Fine-grained Temporal Understanding for Multimodal Video Models},
  author={Cai, Mu and Tan, Reuben and Zhang, Jianrui and Zou, Bocheng and Zhang, Kai and Yao, Feng and Zhu, Fangrui and Gu, Jing and Zhong, Yiwu and Shang, Yuzhang and others},
  journal={arXiv preprint arXiv:2410.10818},
  year={2024}
}

@article{liu2024tempcompass,
  title={Tempcompass: Do video llms really understand videos?},
  author={Liu, Yuanxin and Li, Shicheng and Liu, Yi and Wang, Yuxiang and Ren, Shuhuai and Li, Lei and Chen, Sishuo and Sun, Xu and Hou, Lu},
  journal={arXiv preprint arXiv:2403.00476},
  year={2024}
}

@article{qian2024mia,
  title={Mia-bench: Towards better instruction following evaluation of multimodal llms},
  author={Qian, Yusu and Ye, Hanrong and Fauconnier, Jean-Philippe and Grasch, Peter and Yang, Yinfei and Gan, Zhe},
  journal={arXiv preprint arXiv:2407.01509},
  year={2024}
}

@inproceedings{chiang2023can,
  title={Can large language models be an alternative to human evaluations?},
  author={Chiang, Cheng-Han and Lee, Hung-yi},
  booktitle={ACL},
  year={2023}
}

@inproceedings{mmvet,
  title={Mm-vet: Evaluating large multimodal models for integrated capabilities},
  author={Yu, Weihao and Yang, Zhengyuan and Li, Linjie and Wang, Jianfeng and Lin, Kevin and Liu, Zicheng and Wang, Xinchao and Wang, Lijuan},
  booktitle={ICML},
  year={2024}
}

@inproceedings{li2023semantic,
  title={Semantic-SAM: Segment and Recognize Anything at Any Granularity},
  author={Li, Feng and Zhang, Hao and Sun, Peize and Zou, Xueyan and Liu, Shilong and Yang, Jianwei and Li, Chunyuan and Zhang, Lei and Gao, Jianfeng},
  booktitle={ECCV},
  year={2024}
}

@inproceedings{yolo,
  title={You only look once: Unified, real-time object detection},
  author={Redmon, J},
  booktitle={CVPR},
  year={2016}
}

@article{motsurvey,
  title={Multiple object tracking: A literature review},
  author={Luo, Wenhan and Xing, Junliang and Milan, Anton and Zhang, Xiaoqin and Liu, Wei and Kim, Tae-Kyun},
  journal={AI},
  year={2021},
}

@article{yilmaz2006object,
  title={Object tracking: A survey},
  author={Yilmaz, Alper and Javed, Omar and Shah, Mubarak},
  journal={CSUR},
  year={2006},
}

@inproceedings{meng2023detection,
  title={Detection hub: Unifying object detection datasets via query adaptation on language embedding},
  author={Meng, Lingchen and Dai, Xiyang and Chen, Yinpeng and Zhang, Pengchuan and Chen, Dongdong and Liu, Mengchen and Wang, Jianfeng and Wu, Zuxuan and Yuan, Lu and Jiang, Yu-Gang},
  booktitle={CVPR},
  year={2023}
}

@inproceedings{Meng2023SEGICUT,
  title={SEGIC: Unleashing the Emergent Correspondence for In-Context Segmentation},
  author={Lingchen Meng and Shiyi Lan and Hengduo Li and Jos{\'e} M. {\'A}lvarez and Zuxuan Wu and Yu-Gang Jiang},
  booktitle={ECCV},
  year={2024},
}

@inproceedings{kembhavi2016diagram,
  title={A diagram is worth a dozen images},
  author={Kembhavi, Aniruddha and Salvato, Mike and Kolve, Eric and Seo, Minjoon and Hajishirzi, Hannaneh and Farhadi, Ali},
  booktitle={ECCV},
  year={2016},
}

@inproceedings{yue2024mmmu,
  title={Mmmu: A massive multi-discipline multimodal understanding and reasoning benchmark for expert agi},
  author={Yue, Xiang and Ni, Yuansheng and Zhang, Kai and Zheng, Tianyu and Liu, Ruoqi and Zhang, Ge and Stevens, Samuel and Jiang, Dongfu and Ren, Weiming and Sun, Yuxuan and others},
  booktitle={CVPR},
  year={2024}
}

@inproceedings{yu2019activityqa,
    author = {Yu, Zhou and Xu, Dejing and Yu, Jun and Yu, Ting and Zhao, Zhou and Zhuang, Yueting and Tao, Dacheng},
    title = {ActivityNet-QA: A Dataset for Understanding Complex Web Videos via Question Answering},
    booktitle = {AAAI},
    year = {2019}
}

@article{zhang2024mm1,
  title={Mm1. 5: Methods, analysis \& insights from multimodal llm fine-tuning},
  author={Zhang, Haotian and Gao, Mingfei and Gan, Zhe and Dufter, Philipp and Wenzel, Nina and Huang, Forrest and Shah, Dhruti and Du, Xianzhi and Zhang, Bowen and Li, Yanghao and others},
  journal={arXiv preprint arXiv:2409.20566},
  year={2024}
}

@inproceedings{fu2024objectrelator,
  title={Objectrelator: Enabling cross-view object relation understanding in ego-centric and exo-centric videos},
  author={Fu, Yuqian and Wang, Runze and Fu, Yanwei and Pani Paudel, Danda and Huang, Xuanjing and Van Gool, Luc},
  booktitle={ICCV},
  year={2025}
}

@inproceedings{mckinzie2025mm1,
  title={MM1: methods, analysis and insights from multimodal LLM pre-training},
  author={McKinzie, Brandon and Gan, Zhe and Fauconnier, Jean-Philippe and Dodge, Sam and Zhang, Bowen and Dufter, Philipp and Shah, Dhruti and Du, Xianzhi and Peng, Futang and Belyi, Anton and others},
  booktitle={ECCV},
  year={2025},
}

@inproceedings{li2024densefusion,
  title={Densefusion-1m: Merging vision experts for comprehensive multimodal perception},
  author={Li, Xiaotong and Zhang, Fan and Diao, Haiwen and Wang, Yueze and Wang, Xinlong and Duan, Ling-Yu},
  booktitle={NeurIPS},
  year={2024}
}

@inproceedings{liu2024bench,
  title={Et bench: Towards open-ended event-level video-language understanding},
  author={Liu, Ye and Ma, Zongyang and Qi, Zhongang and Wu, Yang and Shan, Ying and Chen, Chang Wen},
  booktitle={NeurIPS},
  year={2024}
}

@article{yin2024survey,
  title={A survey on multimodal large language models},
  author={Yin, Shukang and Fu, Chaoyou and Zhao, Sirui and Li, Ke and Sun, Xing and Xu, Tong and Chen, Enhong},
  journal={National Science Review},
  year={2024}
}

@article{mmesurvey,
  title={MME-Survey: A Comprehensive Survey on Evaluation of Multimodal LLMs},
  author={Fu, Chaoyou and Zhang, Yi-Fan and Yin, Shukang and Li, Bo and Fang, Xinyu and Zhao, Sirui and Duan, Haodong and Sun, Xing and Liu, Ziwei and Wang, Liang and others},
  journal={arXiv preprint arXiv:2411.15296},
  year={2024}
}

@article{li2024survey,
  title={A survey on benchmarks of multimodal large language models},
  author={Li, Jian and Lu, Weiheng and Fei, Hao and Luo, Meng and Dai, Ming and Xia, Min and Jin, Yizhang and Gan, Zhenye and Qi, Ding and Fu, Chaoyou and others},
  journal={arXiv preprint arXiv:2408.08632},
  year={2024}
}

@inproceedings{wang2024elysium,
  title={Elysium: Exploring object-level perception in videos via mllm},
  author={Wang, Han and Ye, Yongjie and Wang, Yanjie and Nie, Yuxiang and Huang, Can},
  booktitle={ECCV},
  year={2024}
}

@inproceedings{ferret,
  title={Ferret: Refer and ground anything anywhere at any granularity},
  author={You, Haoxuan and Zhang, Haotian and Gan, Zhe and Du, Xianzhi and Zhang, Bowen and Wang, Zirui and Cao, Liangliang and Chang, Shih-Fu and Yang, Yinfei},
  booktitle={ICLR},
  year={2024}
}

@inproceedings{glamm,
  title={Glamm: Pixel grounding large multimodal model},
  author={Rasheed, Hanoona and Maaz, Muhammad and Shaji, Sahal and Shaker, Abdelrahman and Khan, Salman and Cholakkal, Hisham and Anwer, Rao M and Xing, Eric and Yang, Ming-Hsuan and Khan, Fahad S},
  booktitle={CVPR},
  year={2024}
}

@article{li2025benchmark,
  title={Benchmark evaluations, applications, and challenges of large vision language models: A survey},
  author={Li, Zongxia and Wu, Xiyang and Du, Hongyang and Nghiem, Huy and Shi, Guangyao},
  journal={arXiv preprint arXiv:2501.02189},
  year={2025}
}

@inproceedings{refcocog,
author = {Mao, Junhua and Huang, Jonathan and Toshev, Alexander and Camburu, Oana and Yuille, Alan L. and Murphy, Kevin},
title = {Generation and Comprehension of Unambiguous Object Descriptions},
booktitle = {CVPR},
year = {2016}
}

@inproceedings{refcoco,
  title={Referitgame: Referring to objects in photographs of natural scenes},
  author={Kazemzadeh, Sahar and Ordonez, Vicente and Matten, Mark and Berg, Tamara},
  booktitle={EMNLP},
  year={2014}
}

@inproceedings{fu2024cross,
  title={Cross-domain few-shot object detection via enhanced open-set object detector},
  author={Fu, Yuqian and Wang, Yu and Pan, Yixuan and Huai, Lian and Qiu, Xingyu and Shangguan, Zeyu and Liu, Tong and Fu, Yanwei and Van Gool, Luc and Jiang, Xingqun},
  booktitle={ECCV},
  year={2024},
}

@article{openvclip,
  title={Building an open-vocabulary video clip model with better architectures, optimization and data},
  author={Wu, Zuxuan and Weng, Zejia and Peng, Wujian and Yang, Xitong and Li, Ang and Davis, Larry S and Jiang, Yu-Gang},
  journal={TPAMI},
  year={2024},
}

@article{chen2025comp,
  title={Comp: Continual multimodal pre-training for vision foundation models},
  author={Chen, Yitong and Meng, Lingchen and Peng, Wujian and Wu, Zuxuan and Jiang, Yu-Gang},
  journal={ArXiv},
  year={2025}
}

@inproceedings{chen2025comprehensive,
  title={Comprehensive Multi-Modal Prototypes Are Simple and Effective Classifiers for Vast-Vocabulary Object Detection},
  author={Chen, Yitong and Yao, Wenhao and Meng, Lingchen and Wu, Sihong and Wu, Zuxuan and Jiang, Yu-Gang},
  booktitle={AAAI},
  year={2025}
}

\newpage
\appendix
\section*{Appendix}
\begin{itemize}[leftmargin=0.5em]
    \item In \cref{sec:supp_pipeline}, we outline additional implementation details of the GPT-4o-assisted data annotation pipeline.
    \item In \cref{sec:supp_benchmark}, we present further information about the instance understanding benchmark, \ourbenchmark.
    \item In \cref{sec:supp_dataset}, we share more details about the instruction fine-tuning dataset, \oursft.
    \item In \cref{sec:supp_more_discuss}, we provide more discussions on failure cases and real-world applications.
\end{itemize}

\section{Data Annotation Pipeline}
\label{sec:supp_pipeline}
\subsection{Set-of-Marks Visual Prompting}
\label{sec:supp_som}
\begin{figure}[h]
    \centering
    \includegraphics[width=0.9\linewidth]{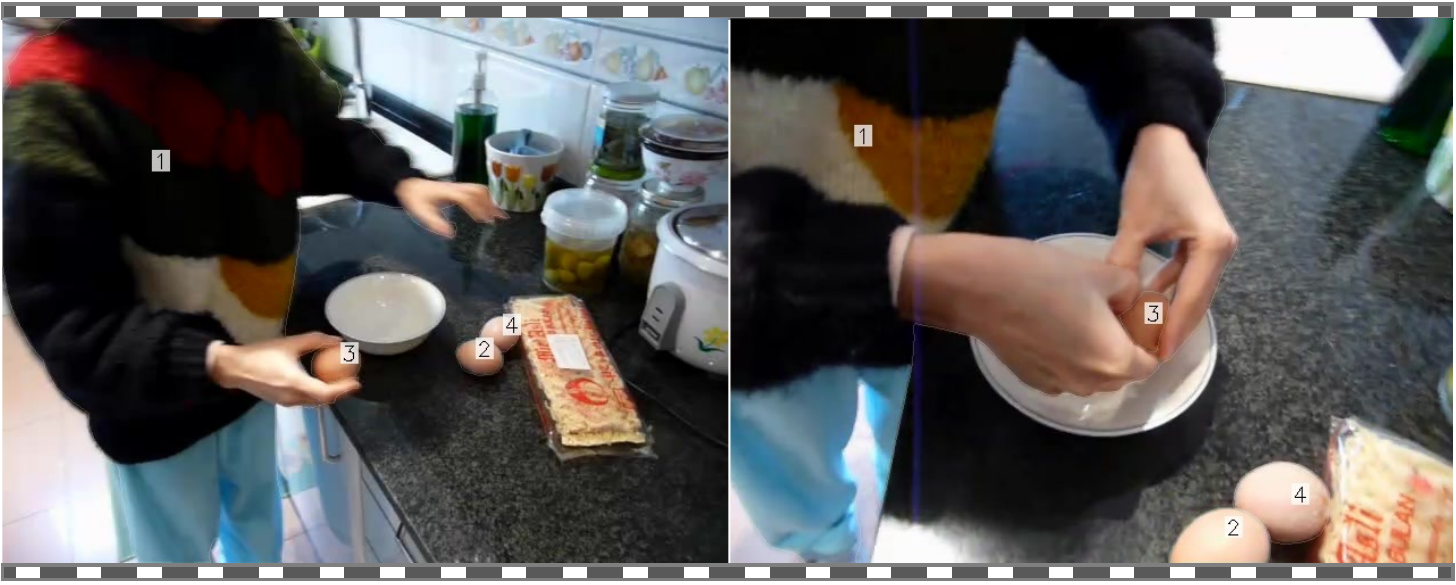}
    \caption{\textbf{Set-of-Marks visual prompting on the original videos.} Each instance is assigned a unique numeric ID, which remains consistent across all frames.}
    \label{fig:supp_som}
\end{figure}

Performing instance-level annotations is challenging, and we adopt the SoM visual prompting technique~\cite{som} to address this. Specifically, as illustrated in \cref{fig:supp_som}, we overlay a numeric ID at the center of each instance and maintain the same ID for a given instance across all frames. This simple augmentation can explicitly guide GPT-4o to focus more effectively on the instances of interest, enabling finer-grained and more accurate annotations. Furthermore, segmentation masks are necessary to calculate the center coordinates of each instance. Details on how these masks are obtained are provided in \cref{sec:supp_dataset_source}.

\subsection{Prompting GPT-4o}
\label{sec:supp_prompt_gpt}
\begin{figure*}
    \centering
    \vspace{2em}
    \includegraphics[width=\linewidth]{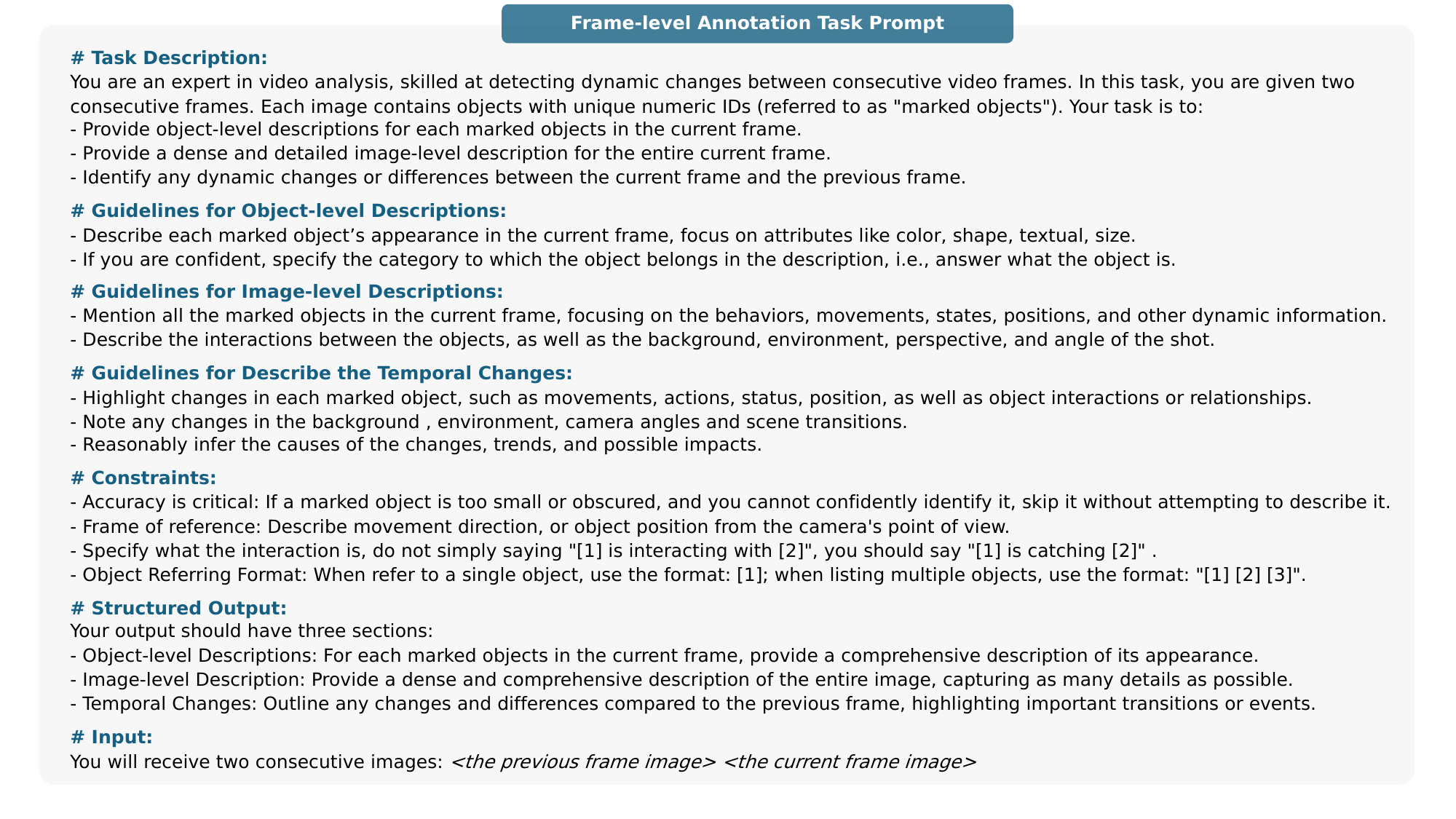}
    \caption{\textbf{Frame-level annotation task prompt}, the \textit{italicized} part are placeholders for the actual inputs.}
    \label{fig:supp_frame_prompt}
\end{figure*}

\begin{figure*}
    \centering
    \includegraphics[width=\linewidth]{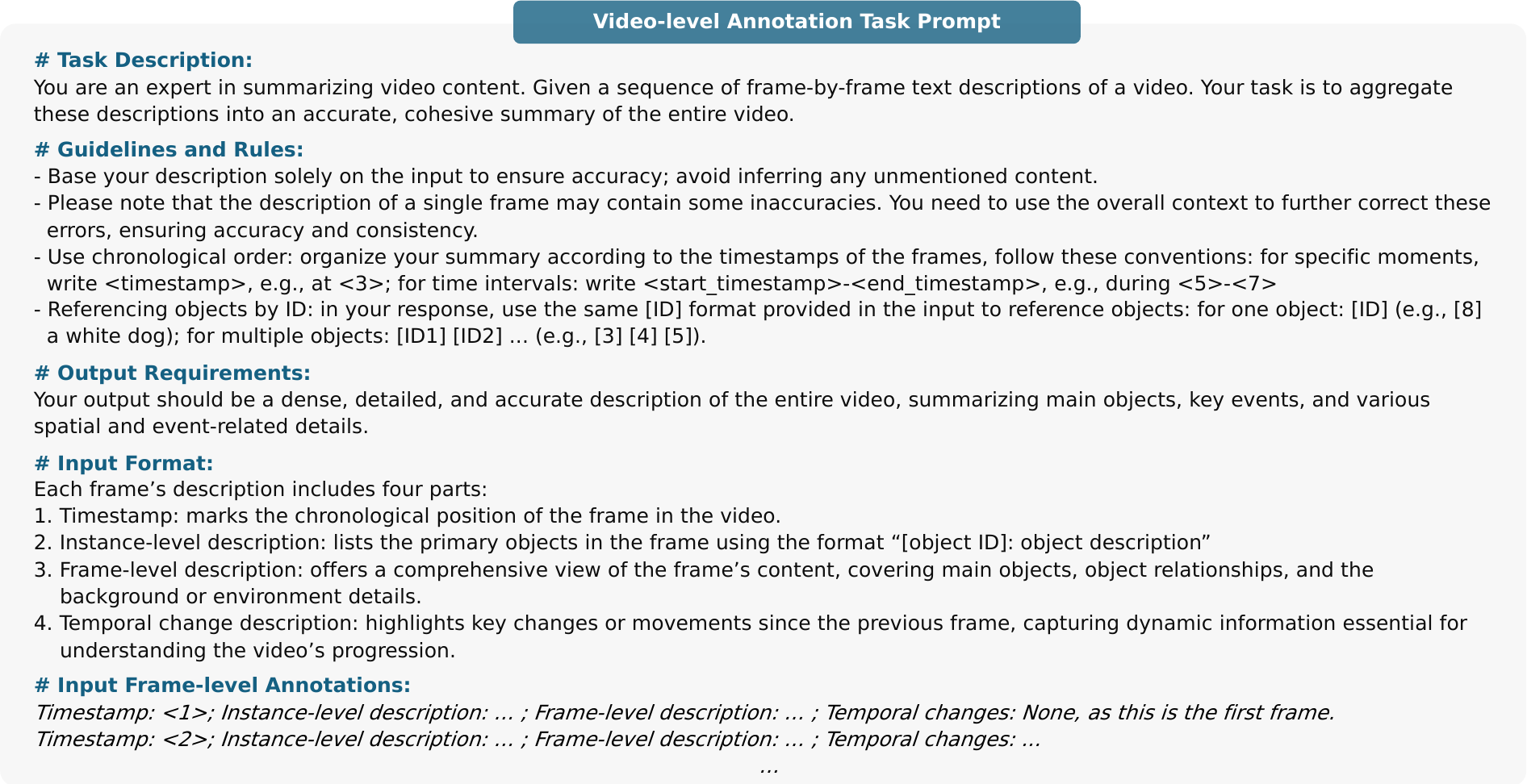}
    \caption{\textbf{Video-level annotation task prompt}, the \textit{italicized} part are placeholders for the actual inputs.}
    \label{fig:supp_video_prompt}
\end{figure*}

\begin{figure*}
    \centering
    \includegraphics[width=\linewidth]{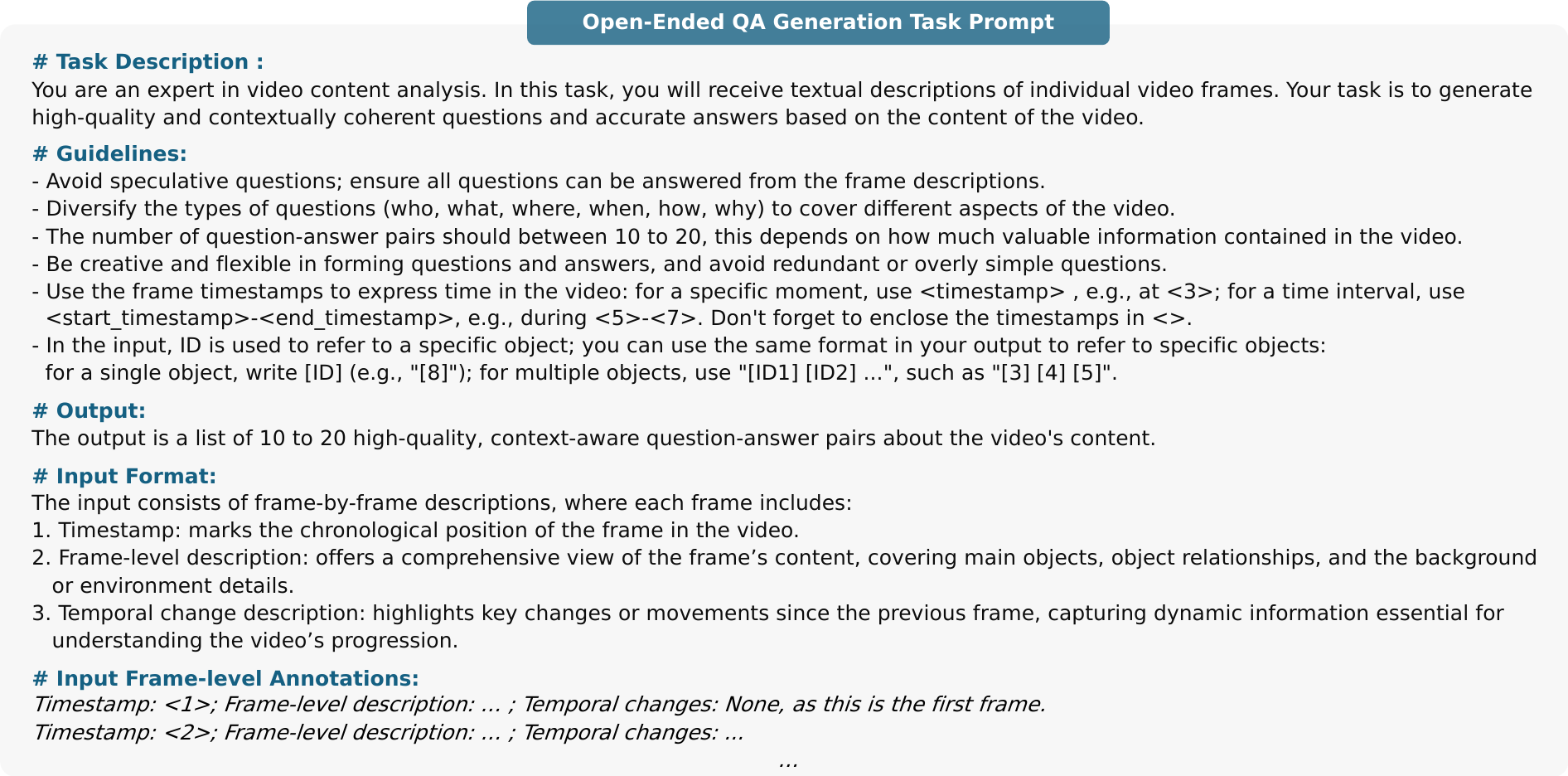}
    \caption{\textbf{Open-ended question-answer pairs generation task prompt}, the \textit{italicized} part are placeholders for the actual inputs.}
    \label{fig:supp_qa_prompt}
\end{figure*}

\fakeparagraph{Task prompt templates.} 
Prompt engineering is crucial for enabling GPT-4o to accomplish specific tasks. In this section, we present the task prompts that we designed to prompt GPT-4o for data annotation: 
\begin{itemize}[leftmargin=1.5em, itemsep=0.1em, topsep=0pt]
    \item The task prompt $P^f$ for frame-level annotation, \cref{fig:supp_frame_prompt}.
    \item The task prompt $P^{vid}$ for video-level annotation, \cref{fig:supp_video_prompt}.
    \item The task prompt $P^{qa}$ for open-ended question-answer pairs generating, \cref{fig:supp_qa_prompt}.
\end{itemize}

\fakeparagraph{GPT-4o API version.}
During the annotation process, we use the GPT-4o-2024-08-06 API and leverage its structured output functionality to facilitate output parsing, enabling the model to respond in a predefined JSON format.

\begin{figure}
    \centering
    \vspace{-1em}
    \includegraphics[width=0.6\linewidth]{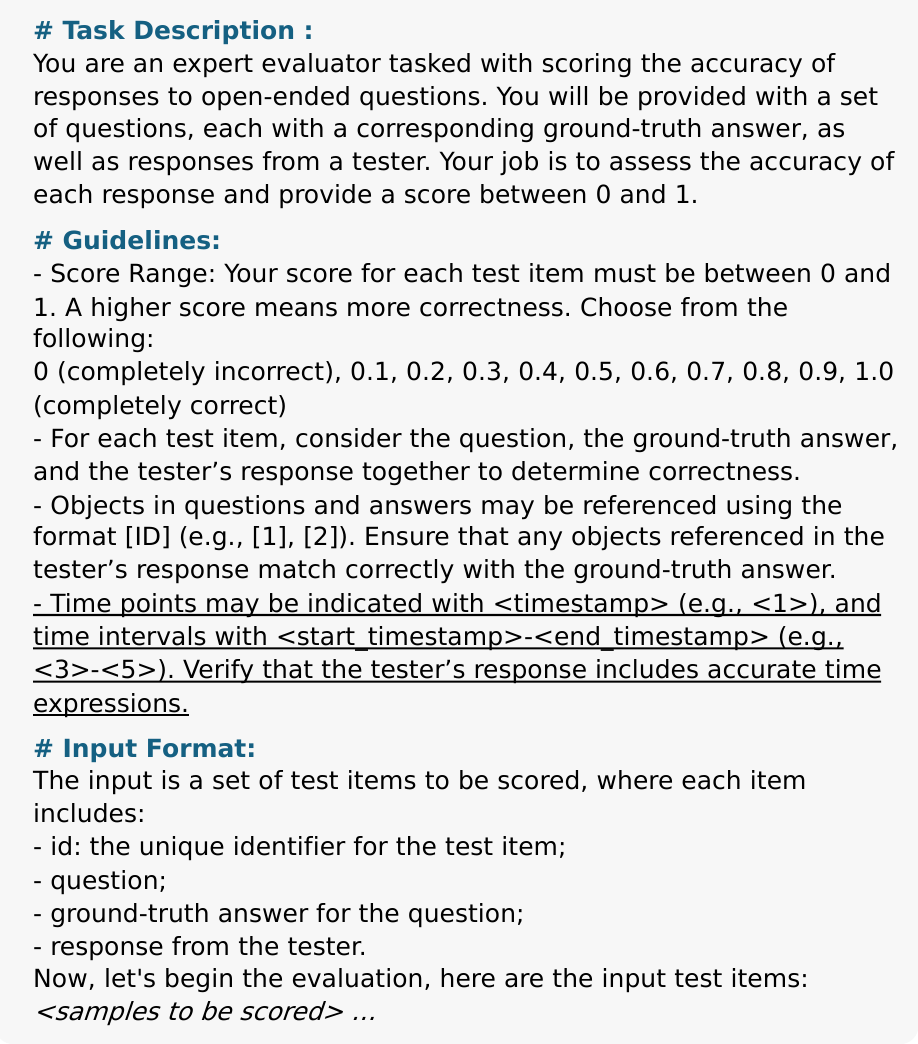}
    \caption{\textbf{GPT-4o-based open-ended question answering correctness assessment.}
    The \underline{underlined} parts in the figure are included only when evaluating the video split, while the \textit{italicized} parts will be replaced by the actual sample for scoring.}
    \vspace{-1em}
    \label{fig:supp_evaluate}
\end{figure}

\begin{figure*}
    \centering
    \vspace{2em}
    \includegraphics[width=\linewidth]{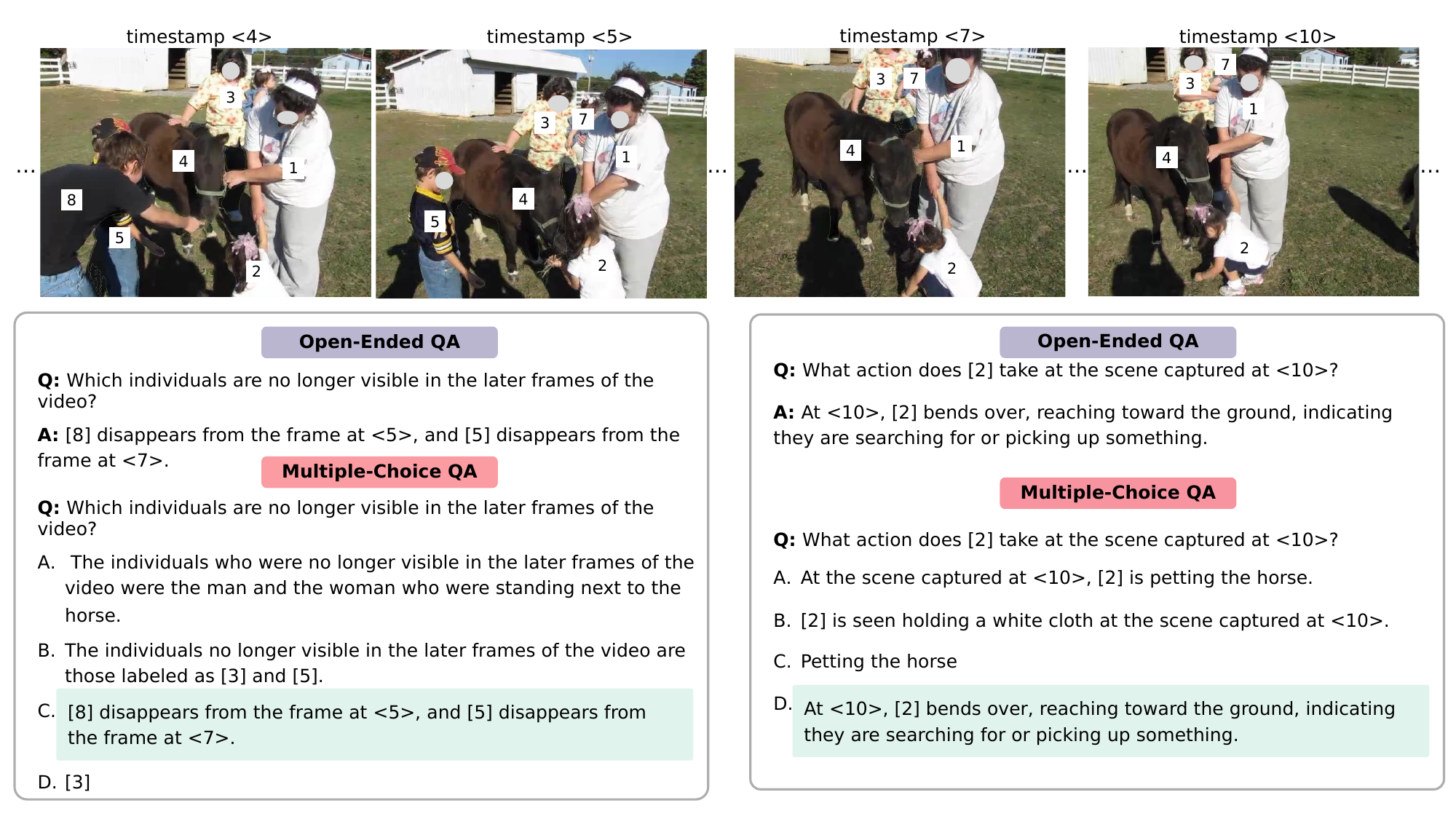}
    \caption{\textbf{A data example from \ourbenchmark.} Each test sample includes both open-ended QA and multiple-choice QA, focusing on specific instances or the relationships and interactions between instances.}
    \label{fig:supp_bench_example}
\end{figure*}

\section{More Details about \ourbenchmark}
\label{sec:supp_benchmark}
\subsection{Negative Options Generation}
\label{sec:supp_bench_create}
We use the ground-truth from open-ended QA as the positive option and additionally craft three negative options, forming a multiple-choice question with four options. 
To create hard negatives, we first have the model answer the open-ended questions and use GPT-4o to score the correctness of the responses. If the score is lower than 0.4, we consider it a difficult negative answer and include it as one of the negative options. Finally, we randomly shuffle the four options to ensure that the correct one appears in each position with equal probability.

\subsection{LLM-based Evaluator for Open-Ended QA}
\label{sec:supp_bench_eval}
Recent studies~\cite{mmvet, chiang2023can} suggest that LLMs can serve as effective evaluators. Building on this, we use GPT-4o to assess the accuracy of open-ended question answering. Specifically, GPT-4o assigns a score between 0 and 1 based on three key factors: the question, the ground-truth answer, and the model prediction. Given that \ourbenchmark prioritizes instance-level understanding, we pay special attention to the accuracy of instance ID references. Furthermore, for the video split of \ourbenchmark, we emphasize the correctness of timestamps to ensure temporal correctness. The task prompt for GPT-4o is illustrated in \cref{fig:supp_evaluate}.

\subsection{Data Example}
To provide a clearer understanding of \ourbenchmark, we present a data example in \cref{fig:supp_bench_example}. 
Each question includes both open-ended and multiple-choice formats, focusing on specific instances or exploring the relationships and interactions between multiple instances. This design highlights the significant distinction from other benchmarks, emphasizing fine-grained understanding at the instance level.

\begin{table}
\centering
\caption{\textbf{Data sources.} We use seven video datasets and one image dataset as our data sources. We show their annotation formats, the splits we used, and the number of samples from each dataset.}
\begin{tabular}{lccc}
\toprule
Dataset Name & Ann. Type & Split & Sample Num. \\ 
\midrule
\multicolumn{4}{c}{\textit{Video Instance Segmentation}} \\ 
BRUST~\cite{brust} & mask  & training & 500 \\
UVO~\cite{uvo} & mask  & training &  5,135 \\
OVIS~\cite{ovis} & mask  & training & 599 \\
LVVIS~\cite{lvvis} & mask  & training &  3,057 \\
YoutubeVIS~\cite{ytvis} & mask  & training & 2,897 \\
\midrule
\multicolumn{4}{c}{\textit{Video Object Tracking}} \\ 
BenSMOT~\cite{bensmot} & box  & training & 2,261  \\
VidOR~\cite{vidor} & box  & training &  6,969 \\
\midrule
\multicolumn{4}{c}{\textit{Image}} \\ 
SA-1B~\cite{ovis} & none  & 1-10 &  51,101  \\
\bottomrule
\end{tabular}%
\label{tab:supp_data_src}
\end{table}

\section{More Details about \oursft}
\label{sec:supp_dataset}
\subsection{Data Collection and Processing}
\label{sec:supp_dataset_source}
\fakeparagraph{Collection.}
We select five instance segmentation datasets and two multi-object tracking datasets as sources of video data. To prevent data leakage, we only used the training splits of these datasets, leaving their test and validation splits untouched. Additionally, we use the SA-1B~\cite{sam} dataset as a source of image data and only utilize the first ten officially provided data splits. For each split, we only use the first 50\% of its images. In total, we collect 21,418 videos and 51,101 images. \cref{tab:supp_data_src} provides detailed statistics on our data sources.

\fakeparagraph{Processing.}
When constructing SoM~\cite{som} visual prompts, we need to obtain the mask annotations for each instance to determine the location of the numeric IDs. For the video instance segmentation datasets~\cite{brust, uvo, ovis, lvvis, ytvis}, the instance masks are already provided and can be used directly. For multi-object tracking datasets~\cite{bensmot, vidor}, we prompt SAM~\cite{sam} with their bounding box annotations to generate instance masks. For images in the SA-1B dataset~\cite{sam}, we employ Semantic-SAM~\cite{li2023semantic} to segment the instances and obtain their masks.

\subsection {Statistics Analysis.}
\label{sec:supp_dataset_statis}

\begin{figure}[t]
  \centering
  \begin{minipage}[t]{0.48\textwidth}
    \centering
    \includegraphics[width=\linewidth]{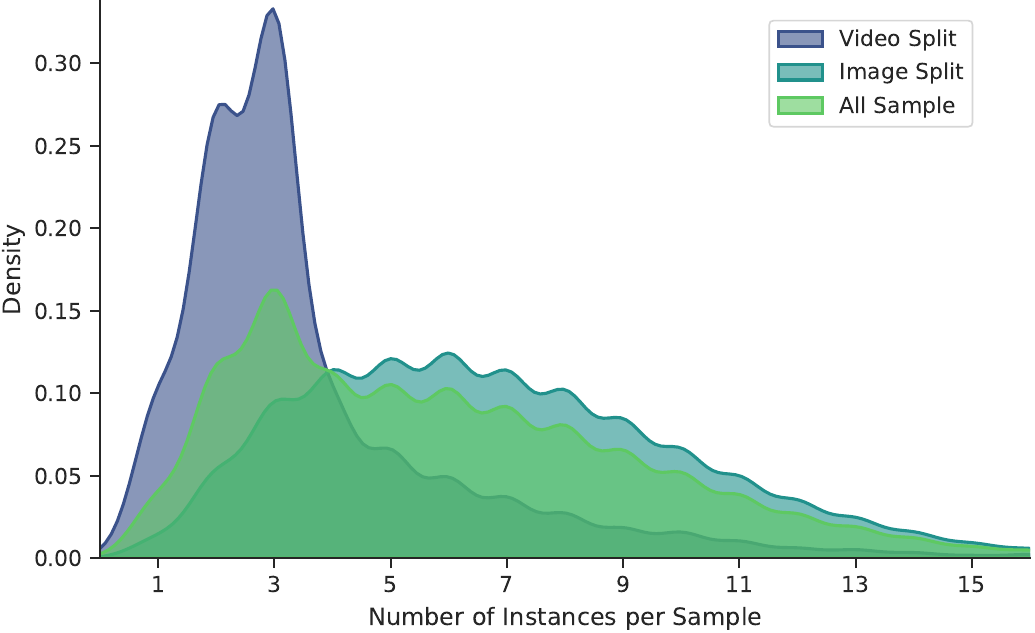}
    \caption{\textbf{The distribution of the number of instances per sample in \oursft.} We separately present the distribution for the video split, image split, and the entire dataset.}
    \label{fig:supp_instance_num}
  \end{minipage}%
  \hfill
  \begin{minipage}[t]{0.48\textwidth}
    \centering
    \includegraphics[width=\linewidth]{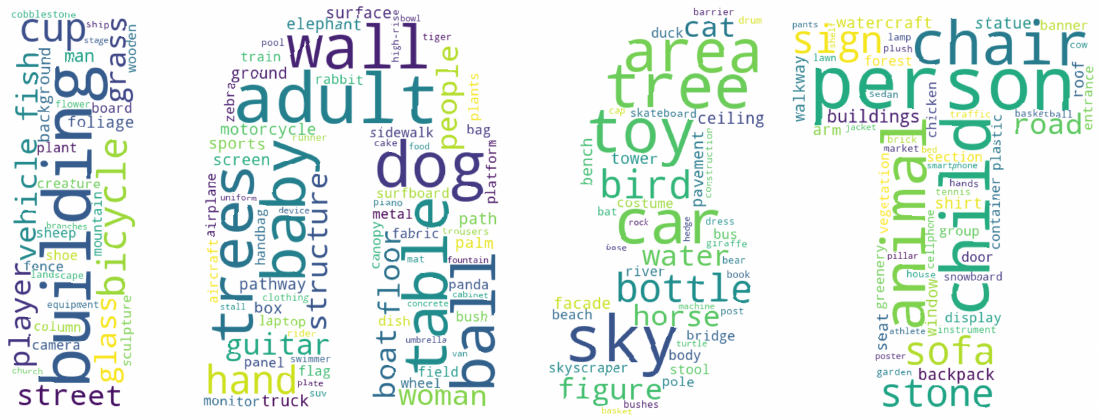}
    \caption{\textbf{Analysis of object categories in \oursft}, which shows a diverse range of types spanning multiple domains.}
    \label{fig:supp_wordcloud}
  \end{minipage}
\end{figure}

\fakeparagraph{Number of instances.}
The key characteristic of \oursft is its specific focus on individual instances in images and videos, which provides a more fine-grained description of the visual inputs. We visualize the distribution of the number of instances in each sample in~\cref{fig:supp_instance_num}.
For the video split, each sample has an average of 3.7 instances, with a total of 79,709 instances.
For the image split, each sample contains an average of 6.9 instances, totaling 351,495 instances.
Across the entire dataset, each sample includes an average of 5.9 instances, adding up to 431,204 instances in total. 
We measure the scene complicity by the number of instances in each sample. Specifically, 31\% of the samples contain $\leq$ 3 instances (simple), 39\% have between 3 to 8 instances (medium), and the remaining 30\% contain $\geq$ 8 instances (hard).

\fakeparagraph{Dataset diversity.}
We visualize the object categories in \oursft in~\cref{fig:supp_wordcloud}, highlighting its diverse range. 
The objects include humans, animals, plants, vehicles, landmarks, \etc, covering domains like daily life, egocentric perspectives, sports, transportation, \etc.
The rich diversity of \oursft ensures its applicability to real-world scenarios and enhances its transferability to different domains. 

\fakeparagraph{Text captions.}
\oursft contains multi-level textual descriptions of visual content, covering instances, frames, temporal changes, and video-level annotations. We conduct statistical analysis on these text annotations, including the number of each type of text, and their average length. 
As shown in~\cref{tab:supp_data_statis}, the average length of \oursft is 49.1 words per caption, with video-level averaging 323.2 words, highlighting its richness of details. 
We also present the results of lexical analysis in~\cref{tab:supp_data_statis}. The instance-level captions contain a rich variety of nouns and adjectives, indicating that they primarily describe the objects' categories and attributes. The captions of temporal changes include a high volume of verbs and adverbs, suggesting that they capture dynamic information.

\fakeparagraph{Human evaluation of data quality}
We invited three volunteers to rate each sample on a scale from 1 to 5, with higher scores indicating better quality. \cref{tab:human_eval} presents the scores of different types of annotations, along with the average time spent by each volunteer to evaluate each sample. The average score across all types is 4.49$_{\pm{0.05}}$, indicating that the data in \oursft is of satisfactory quality.

\begin{table}
\centering
\caption{\textbf{Statistical and lexical analysis of \oursft.} We present the results for each annotation level as well as the entire dataset.}
\resizebox{\linewidth}{!}{%
\begin{tabular}{l|c|ccc|ccccc}
\toprule
Caption Type & \#Caption & \#Char./Cap. & \#Word/Cap. & \#Sen./Cap. & Nouns & Adj. & Adv. & Verb. & Prep. \\ 
\midrule
Instance-level  & 836,524 & 102.1  & 24.3  & 1.5  & \textbf{26.5\%} & \textbf{13.3\%} & 2.3\% & 12.3\% & 10.7\% \\
Frame-level     & 207,662 & 458.0  & 106.5 & 5.7  & 25.2\% & 10.5\% & 2.6\% & 14.9\% & 11.5\% \\
Temporal-change & 135,143 & 306.6  & 67.7  & 3.7  & 21.2\% & 10.0\% &\textbf{ 6.0\%} & \textbf{16.4\%} & 10.8\% \\
Video-level     & 21,372  & 1441.8 & 342.2 & 14.3 & 24.8\% & 10.6\% & 3.6\% & 13.2\% & \textbf{11.8\%} \\
\rowcolor{gray!10}
All & 1,200,701 & 210.5   & 49.1   & 2.7   & 25.0\% & 11.4\% & 3.1\% & 14.0\% & 11.1\% \\
\bottomrule
\end{tabular}%
}
\label{tab:supp_data_statis}
\end{table}

\begin{table}[t]
\centering
\caption{Human evaluation on the quality of \oursft.}
\resizebox{0.9\linewidth}{!}{%
\begin{tabular}{lccccc} \toprule
& Instance Caption & Image Caption & Temporal Caption & Video Caption & QA Pairs \\ \cmidrule(lr){1-6}
Score ($\uparrow$) & 4.66$_{\pm{0.12}}$ & 4.68$_{\pm{0.02}}$ &  
4.48$_{\pm{0.05}}$ & 4.34$_{\pm{0.18}}$ & 4.31$_{\pm{0.11}}$ \\ 
Time (s) & 7.3 &  12.4  &  11.9  & 31.0 & 10.6 \\ \bottomrule
\end{tabular}%
}
\label{tab:human_eval}
\end{table}

\subsection{Data example.}
\label{sec:supp_dataset_exp}
In this section, we provide a complete video data sample from \oursft to offer a clearer understanding of its content and format. 
In all annotations, we use the format [ID] to refer to instances and $<$timestamp$>$ to refer to timestamps. We present the frame-level annotations in \cref{tab:supp_dataset_example_1}. We can see that each frame-level annotation $Y^f$ consists of three parts: instance-level descriptions $y^{ins}$, image-level descriptions $y^{img}$, and temporal differences $y_{dif}$. Additionally, each video is accompanied by a series of open-ended question-answer pairs $Y^{qa}$, most of which center on specific instances or their relationships, as illustrated in \cref{tab:supp_dataset_example_qa}. Furthermore, we generate a dense video-level caption $Y^{vid}$ summarizing the entire video in chronological order, as shown in \cref{tab:supp_dataset_example_vid_cap}.

\section{More discussions.}
\label{sec:supp_more_discuss}
\subsection{Failure cases.}
We manually inspect the dataset and model to identify the failure cases. We find that occasional failures occur in scenarios where instances are severely occluded, the image is blurry, or instances are excessively small or crowded. These challenges are common among LMMs, and future research can further investigate them.

\subsection{Real-world applications.}
In real-world applications, users can interactively prompt models like SAM2~\cite{ravi2024sam} to automatically track instances of interest and generate SoMs. Additionally, our model also supports inputs without SoMs, allowing users to specify particular instances using textual descriptions. 
In the first scenario, our \ourmethod introduces only a marginal overhead for generating SoMs, while in the second case, it incurs no extra cost compared to the base model.

\begin{table}
\centering
\caption{\textbf{\oursft frame-level annotations}. For the ease of visualization, we only demonstrate the first three frames. Please zoom in to view the instance ID labels.}
\label{tab:supp_dataset_example_1}
\smaller[1]
\renewcommand{\arraystretch}{1.2} %

\begin{tabular}{m{3cm}m{4cm}m{4cm}m{3.5cm}}
\toprule
\textbf{Frame} &\textbf{ Instance-level captions} & \textbf{Image-level captions} &\textbf{ Temporal differences} \\
\midrule
\includegraphics[width=0.15\textwidth]{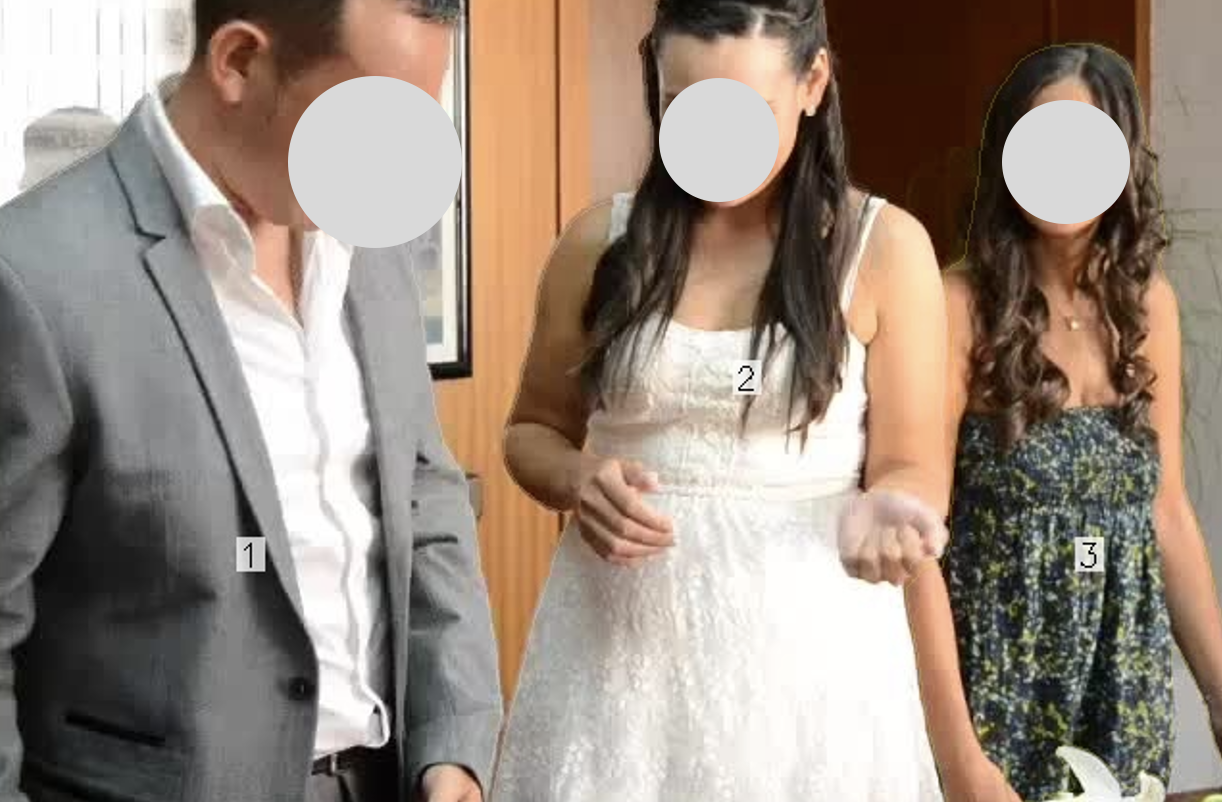}
\centering timestamp$<$1$>$
&
1: Wearing a light gray suit with a white shirt, standing indoors. 2: Wearing a sleeveless white lace dress, holding an object in the hand. 3: Wearing a dark floral-patterned dress with long wavy hair.
&
[1] [2] [3] are standing closely together in an indoor setting. [1] is on the left side wearing a formal, light gray suit with a white shirt. [2], in the middle, is wearing a sleeveless white lace dress, holding something in their hand. [3] is on the right side in a dark floral-patterned dress with long, wavy hair. They appear to be in a room with wooden paneling and some framed art on the wall.
null
&

\\
\includegraphics[width=0.15\textwidth]{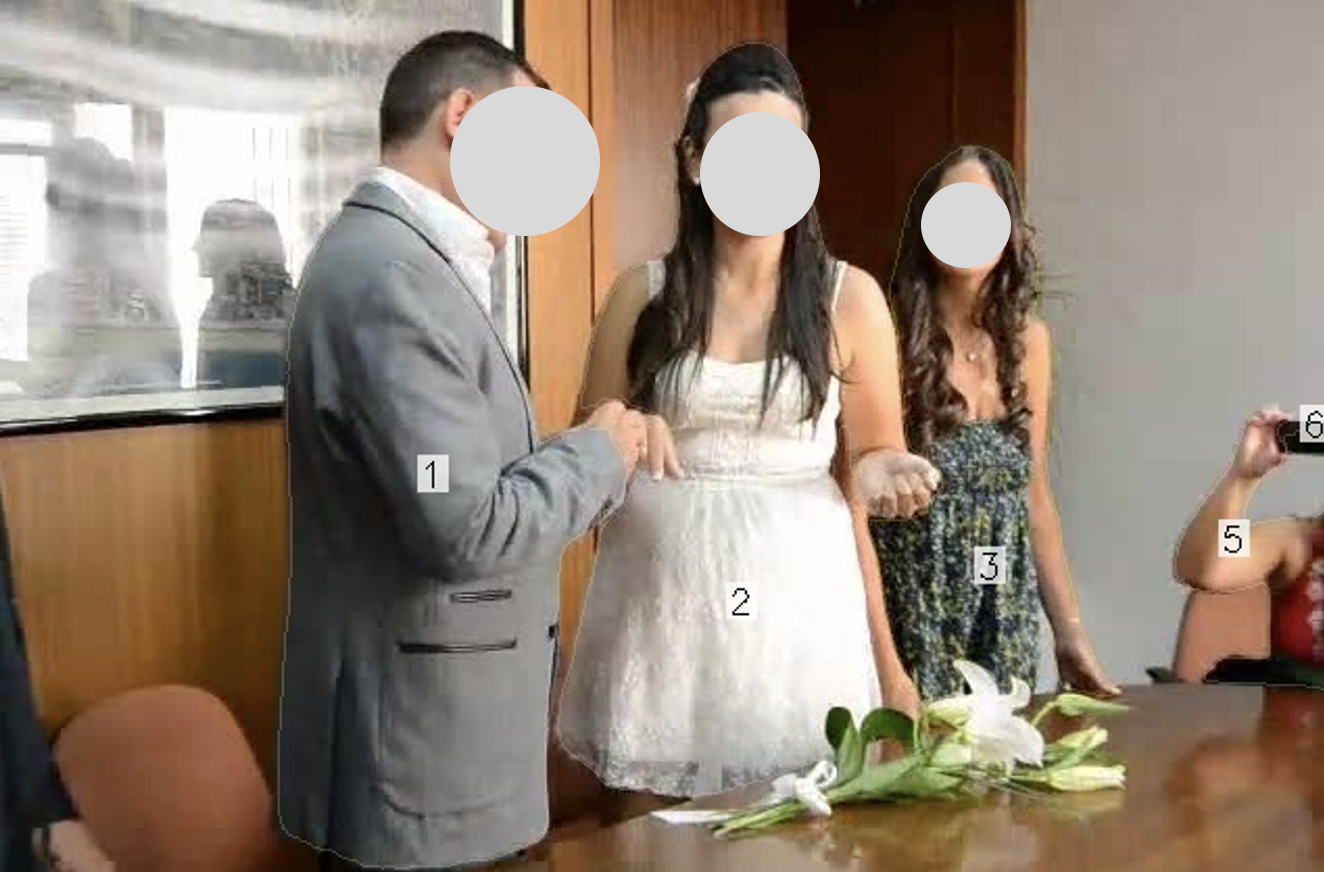} 
\centering timestamp$<$2$>$
&
1: A person wearing a gray suit with a white shirt, short hair. 2: A person in a white, sleeveless dress with long dark hair. 3: A person wearing a dark floral dress with long dark hair. 5: A person wearing red, partially visible in the background. 6: A small black cellphone held in a hand.
&
The scene appears to be in an office setting with a wooden table at the foreground. [1] is standing to the left, facing [2], and appears to be holding [2]'s finger or hand. [2] stands slightly to the right, returning focus with [1]. [3] is to the right of [2], slightly in the background, smiling and looking forward. A bouquet of white flowers lies on the table near [2]. [5] is partially visible in the background on the right, seated and wearing red. [6] is a cellphone held by [5]. Background shows a wooden wall and a reflection in a window.
&
[1] has moved closer to [2] and is now in contact with [2]'s hand. [2] has turned slightly towards [1] compared to the previous frame. [3] remains in a similar position, but the expression suggests more engagement with the scene. [5] and [6] have appeared in the frame; [5] is visible in the background holding [6]. The table with a bouquet of flowers is now visible, indicating a shift in camera angle slightly to include more of the right side of the room.
\\
\includegraphics[width=0.15\textwidth]{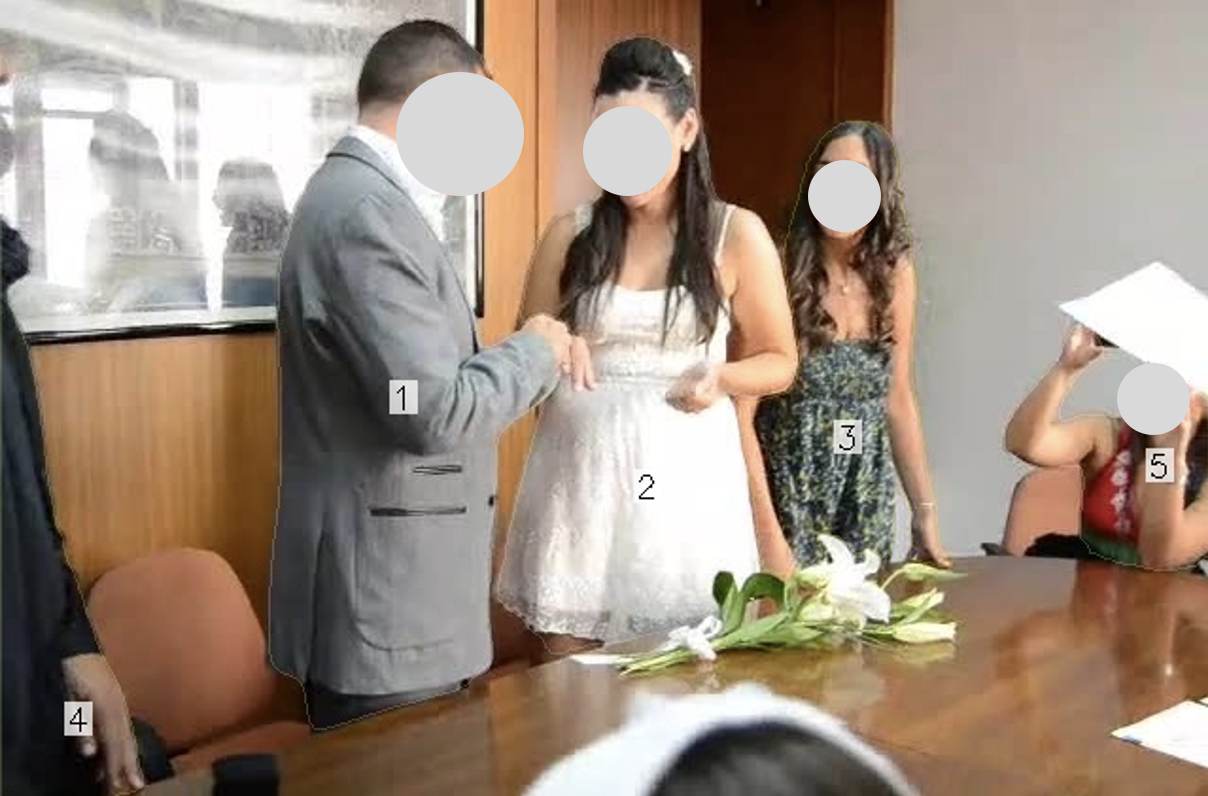}
\centering timestamp$<$3$>$
&
1: Wearing a grey suit, standing beside [2] and slightly turned towards them. 2: Wearing a white, sleeveless dress with floral textures. Holding a bouquet of white flowers. 3: Wearing a dark patterned dress, standing slightly behind [2]. 4: Partially visible, wearing dark clothing, located at the edge of the left side of the frame. 5: Seated, wearing a red outfit. Holding a white object above their head, possibly obscuring their face.
&  
The scene shows [1] [2] [3] near a wooden conference table in a professional setting, possibly an office. [1] wears a grey suit and is standing to the left, engaged with [2] who is wearing a white dress and holding flowers. [3], who is in a patterned dress, stands closely behind [2]. The newly appeared [4] is seated to the far left, partially visible at the edge of the frame. [5] is seated on the right side, holding an object above their head, possibly obscuring their face. The room has wooden walls and a framed picture hanging on the wall.
& 
Object [5] has lifted an object above their head, possibly a piece of paper. Object [4] has appeared in the scene, seated on the left side of the frame, which was not visible earlier. The positions of objects [1], [2], and [3] remain unchanged, as does the background and setting of the room. Overall, no significant movement is noticed in terms of camera angle or position for objects [1] [2] [3].
\\

\bottomrule
\end{tabular}
\end{table}

\begin{table}
\centering
\caption{\textbf{\oursft Open-ended question-answer pairs.}}
\renewcommand\arraystretch{1.6}
\begin{tabular}{p{0.45\textwidth} p{0.45\textwidth}}
\toprule
\multicolumn{1}{c}{\textbf{Question}} & \multicolumn{1}{c}{\textbf{Answer}} \\ 
\midrule
What change occurs with [1]'s expression between $<$10$>$ and the previous frame? & [1] changes from smiling to a neutral expression. \\
What activity are [1] and [2] involved in at $<$11$>$? & [1] and [2] are engaged in a kiss. \\
What is the overall mood during $<$11$>$ as suggested by [3]'s actions? & A celebratory or joyous event. \\
What interaction occurs between [1] and [2] at $<$5$>$? & [1] holds [2]'s hand, suggesting an intimate gesture or exchange, likely a ring. \\
Who joins [1] and [2] in the frame at $<$7$>$? & [4] appears in the frame, joining [1] and [2]. \\
What changes in the group's composition between $<$7$>$ and $<$8$>$? & [3] reappears, and [4] is no longer visible. \\
What element is seen throughout the frames $<$1$>$ to $<$12$>$? &The scene is in an indoor setting with wooden paneling and framed art. \\
What type of event is likely taking place based on the atmosphere in $<$4$>$ and $<$6$>$? & A formal event, possibly a wedding or official gathering. \\
What new elements are introduced in the scene at $<$2$>$? & [5] holds a cellphone in the background, partially visible. \\
What is the mood and lighting like at $<$6$>$? & The mood is formal and celebratory, with bright lighting enhancing this atmosphere. \\
What new background element appears at $<$7$>$? & There is a map or blueprint on the wall. \\
What is notable about [5]'s actions at $<$3$>$? & [5] is lifting an object above their head, possibly a piece of paper. \\ 
What is the setting like in $<$3$>$? & The group is gathered near a wooden conference table in a formal setting. \\
How are [1] and [2] interacting at $<$8$>$? & They are engaged in conversation or communication, indicated by body language and focus. \\
What does [1]'s expression suggest at $<$12$>$? & [1] speaks or smiles, suggesting engagement with [2] or others. \\
What shift occurs in the focus of the camera between $<$5$>$ and $<$6$>$? &The camera focuses more on individuals standing together, reducing focus on the foreground objects. \\
What are [3] and [4] doing at $<$9$>$? & They clapping their hands in celebration. \\
What decorative element is visible at $<$2$>$? & A bouquet of flowers lies on the table near [2]. \\
How has the posture of [1] and [2] changed by $<$6$>$? & [1] and [2] face slightly outward, suggesting a pose for a photograph or audience. \\
What overall physical change occurs between [1] and [2] from $<$10$>$ to $<$11$>$? & There's a noticeable increase in their physical interaction, enhancing emotional engagement. \\
\bottomrule
\end{tabular}
\label{tab:supp_dataset_example_qa}
\end{table}

\begin{table}[t]
\centering
\caption{\textbf{\oursft video-level caption.}}
\renewcommand\arraystretch{1.6}
\begin{tabular}{p{0.95\textwidth}}
\toprule
\multicolumn{1}{c}{\textbf{Video-level caption}} \\ 
\midrule
The video appears to document a formal or celebratory event indoors, possibly a ceremony such as a wedding or official gathering, occurring in a room with wooden paneling and art or framed pictures on the wall. At the beginning, during $<$1$>$, [1] is wearing a light gray suit and stands with [2] in a sleeveless white lace dress, and [3] in a dark floral-patterned dress. The three are close together, suggesting an intimate or focused setting. The progression between $<$2$>$ and $<$3$>$ involves subtle changes in posture and interaction. [1] moves closer to [2], appearing to hold hands or engage in an exchange, possibly involving a ring, as indicated by a bouquet of flowers. [3] remains supportive and smiling, while [5], in red, momentarily holds an object above their head, before disappearing from view by $<$4$>$.In frames $<$5$>$ to $<$7$>$, [1] and [2] maintain a close interaction, suggestive of a significant moment such as an exchange of vows or rings. They are closely observed by [3], who stands smiling nearby, while [1] and [2] occasionally adjust their positions, facing each other initially and then turning outward, which may signal transitioning from an intimate moment to posing for a photo. By $<$7$>$, [4] joins, dressed in darker attire, emphasizing the formal setting as [3] is no longer visible. Through $<$8$>$ and $<$9$>$, the group dynamics change slightly with the absence of [4] and [3] entering the scene again. [1] and [2] appear to engage in a warm interaction as [3] supports them, clapping, alongside the visible hands of [4] indicating applause, marking a cheerful tone. Finally, during $<$10$>$ to $<$12$>$, the focus shifts as [1] and [2] first engage in a kiss, underscoring an intimate conclusion to their ceremony. They later stand apart slightly at the center, with [1] smiling or speaking, and [2] leaning towards [1] suggestively content. Throughout, the consistent joyous mood is accentuated by [3]'s ongoing clapping and expression of joy, emphasizing shared celebration and approval from the audience captured. \\
\bottomrule
\end{tabular}
\label{tab:supp_dataset_example_vid_cap}
\end{table}

\section{Limitations and broader impacts.}
\label{sec:limitations_supp}
\fakeparagraph{Limitations.}
Our current experiments are conducted on 7B and 1.5B models due to the computation cost. Moreover, our current data pipeline is automated but constrained by the overhead of GPT-4o. We can further scale the model size and scale the dataset using a model-in-the-loop approach and improve the model through multi-round instruction tuning with self-synthesized data. We leave this direction for future work.

\fakeparagraph{Broader impacts.}
This paper proposes an enhancement of instance-level understanding capabilities in large multimodal models, enabling them to better assist users by answering questions about the content of interest. However, similar to existing large multimodal models, this approach also faces potential risks, such as issues related to fairness and bias. Future work can address this issue through approaches such as data filtering and validation.

\end{document}